\documentclass{article} 
\usepackage{preprint,times}

\usepackage{amsmath,amsfonts,bm}

\def\eqref#1{equation~\ref{#1}}

\def\1{\bm{1}}

\DeclareMathAlphabet{\mathsfit}{\encodingdefault}{\sfdefault}{m}{sl}
\SetMathAlphabet{\mathsfit}{bold}{\encodingdefault}{\sfdefault}{bx}{n}

\usepackage{booktabs}
\usepackage{hyperref}
\usepackage{url}
\usepackage{graphicx}
\usepackage{listings}
\usepackage{amssymb}
\usepackage{placeins}

\fancypagestyle{institutional-first-page}{%
  \fancyhead{}%
  \fancyhead[L]{\includegraphics[width=1.75in]{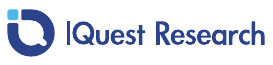}}%
  \fancyhead[R]{\includegraphics[height=0.35in]{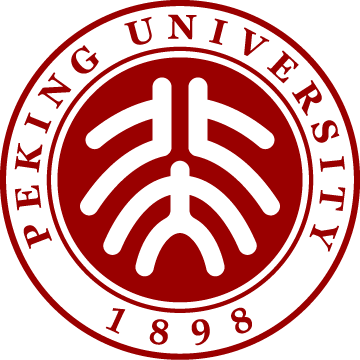}}%
  \renewcommand{\headrulewidth}{0.4pt}%
}

\lstdefinelanguage{Lean}{
  morekeywords={
    theorem, lemma, example, def, by, induction, induction',
    with, have, exact, apply, intro, intros, rw, simp, simp_all,
    ring, ring_nf, abel, abel_nf, aesop, grind, cases, rcases,
    rfl, clear, revert, repeat, symm, try, positivity
  },
  sensitive=true,
  morecomment=[l]{--},
  morestring=[b]"
}

\lstdefinestyle{leancode}{
  language=Lean,
  basicstyle=\ttfamily\scriptsize,
  columns=fullflexible,
  keepspaces=true,
  breaklines=true,
  breakatwhitespace=false,
  showstringspaces=false,
  aboveskip=5pt,
  belowskip=5pt,
  numbers=left,
  numberstyle=\tiny,
  numbersep=5pt,
  xleftmargin=1.25em,
  frame=single,
  framerule=0.4pt,
  framesep=3pt,
  captionpos=b,
  escapeinside={(*@}{@*)},
  literate=
    {ℕ}{{$\mathbb{N}$}}1
    {ℤ}{{$\mathbb{Z}$}}1
    {ℚ}{{$\mathbb{Q}$}}1
    {ℝ}{{$\mathbb{R}$}}1
    {→}{{$\to$}}1
    {←}{{$\leftarrow$}}1
    {↔}{{$\leftrightarrow$}}1
    {≤}{{$\leq$}}1
    {≥}{{$\geq$}}1
    {≠}{{$\neq$}}1
    {∈}{{$\in$}}1
    {ι}{{$\iota$}}1
    {∉}{{$\notin$}}1
    {∑}{{$\sum$}}1
    {∏}{{$\prod$}}1
    {⊢}{{$\vdash$}}1
    {‹}{{$\langle$}}1
    {›}{{$\rangle$}}1
    {λ}{{$\lambda$}}1
    {∀}{{$\forall$}}1
    {∃}{{$\exists$}}1
    {∧}{{$\land$}}1
    {∨}{{$\lor$}}1
    {✝}{{$\dagger$}}1
    {α}{{$\alpha$}}1
    {φ}{{$\phi$}}1
    {Σ}{{$\Sigma$}}1
    {⇑}{{$\Uparrow$}}1
    {≃}{{$\simeq$}}1
    {⁻¹}{{$^{-1}$}}2
    {·}{{$\cdot$}}1
    {⁻}{{$^{-}$}}1
    {¹}{{$^{1}$}}1
    {₀}{{$ _0$}}1
    {⟨}{{$\langle$}}1
    {⟩}{{$\rangle$}}1
}

\title{Reap: Uncovering Monte Carlo proof search with a lightweight model\thanks{The Reap framework is open-source at \url{https://github.com/frenzymath/reap}, and Reaper-1 model weights are available at \url{https://huggingface.co/FrenzyMath/Reaper-1}. The training pipeline code is being prepared for public release.}}

\author{%
  \parbox{\dimexpr\textwidth-2\tabcolsep\relax}{\centering
  Yutong Wang\textsuperscript{1} \quad Xuanzhi Ren\textsuperscript{1} \quad Tianyi Xu\textsuperscript{1} \quad Ruitong Liu\textsuperscript{1} \\[2pt]
  Jingda Xu\textsuperscript{2} \quad Peihao Wu\textsuperscript{2} \quad Bryan Dai\textsuperscript{2} \quad Bin Dong\textsuperscript{3,4,5,6} \\[6pt]
  \normalfont\small
  \textsuperscript{1}Peking University \\
  \textsuperscript{2}IQuest Research \\
  \textsuperscript{3}Beijing International Center for Mathematical Research and the New Cornerstone Science Laboratory, Peking University \\
  \textsuperscript{4}Center for Machine Learning Research, Peking University \\
  \textsuperscript{5}Center for Intelligent Computing, Great Bay Institute for Advanced Study, Great Bay University \\
  \textsuperscript{6}Zhongguancun Academy
  }
}

\begin{document}

\vspace*{6pt}
\maketitle
\thispagestyle{institutional-first-page}
\fancyhead{}
\renewcommand{\headrulewidth}{0pt}

\begin{abstract}
We present \textbf{Reap}, an open framework for learning and analyzing
formal proof search in Lean~4. Reap implements policy--value-guided
Monte Carlo Tree Search directly in Lean, sharing proof-state transitions,
AND--OR search, and replay verification between inference and training
rollouts. Structured action logs expose how neural predictions and tactic
execution interact during search. Using this framework, we train
\textbf{Reaper-1}, a 1.7B-parameter policy--value model, through continued
pre-training, supervised tactic learning, and iterative search-based
reinforcement learning. Using a separate premise selector, the final system
solves 79.9\% of miniF2F-test at pass@32, establishing state-of-the-art performance among open provers of comparable size while remaining competitive with substantially larger models.
Beyond improved solve rates, our results suggest that RL strengthens the prover’s ability to organize proofs around useful intermediate results, favoring substantial advances over repeated local adjustments. This more directed proof construction is accompanied by increasingly elaborate tactic programs: fewer explicit search decisions are needed, while more work is performed within each decision. The resulting trade-off between search effort and tactic execution cost provides insight into how learning changes the prover’s problem-solving strategy.

\end{abstract}

\section{Introduction}

Formal theorem proving provides a reliable testbed for studying machine reasoning: proof assistants expose explicit proof states and verify completed proofs with a trusted kernel, grounding reasoning in a precisely defined environment. Recent whole-proof systems, such as Kimina-Prover, DeepSeek-Prover-V2, and Seed-Prover~\citep{kiminaprover,deepseekproverv2,seedprover}, have achieved strong mathematical performance with increasingly large models and elaborate inference pipelines, while step-level reinforcement learning has substantially improved neural theorem provers~\citep{htps,alphaproof}. These advances raise a complementary question: beyond solving more theorems, how does training change the mechanism and procedure of proof search? Earlier systems~\citep{htps,polu2020gptf} combined relatively small models with inspectable search environments, whereas more recent high-performing provers~\citep{alphaproof,bfsproverv2,internlm25stepprover} can be difficult to study independently because of limited implementation access or substantial computational requirements. This motivates our central goal: combining strong theorem-proving performance with an accessible and observable search framework.

To this end, we introduce \textbf{Reap}, an open framework that implements policy--value-guided Monte Carlo Tree Search (MCTS) directly as a Lean~4 tactic. Reap runs the semantics-critical search loop inside Lean, supporting native proof-state transitions, AND--OR search, runtime budget control, proof reconstruction, and replay verification, while external neural services provide policy and value predictions. The same implementation serves both as an inference-time proof-search tactic and as a rollout engine for reinforcement learning, providing a consistent interface between training and evaluation. Reap also records structured search traces, making the interaction between neural predictions, tactic execution, and search directly observable. Together with the distributed \textbf{Scythe-rl} pipeline for coordinating rollouts, training, and data management, this design provides a lightweight environment for developing and analyzing search-based formal provers.

\begin{figure}[!htbp]
\begin{center}
\includegraphics[width=0.88\linewidth]{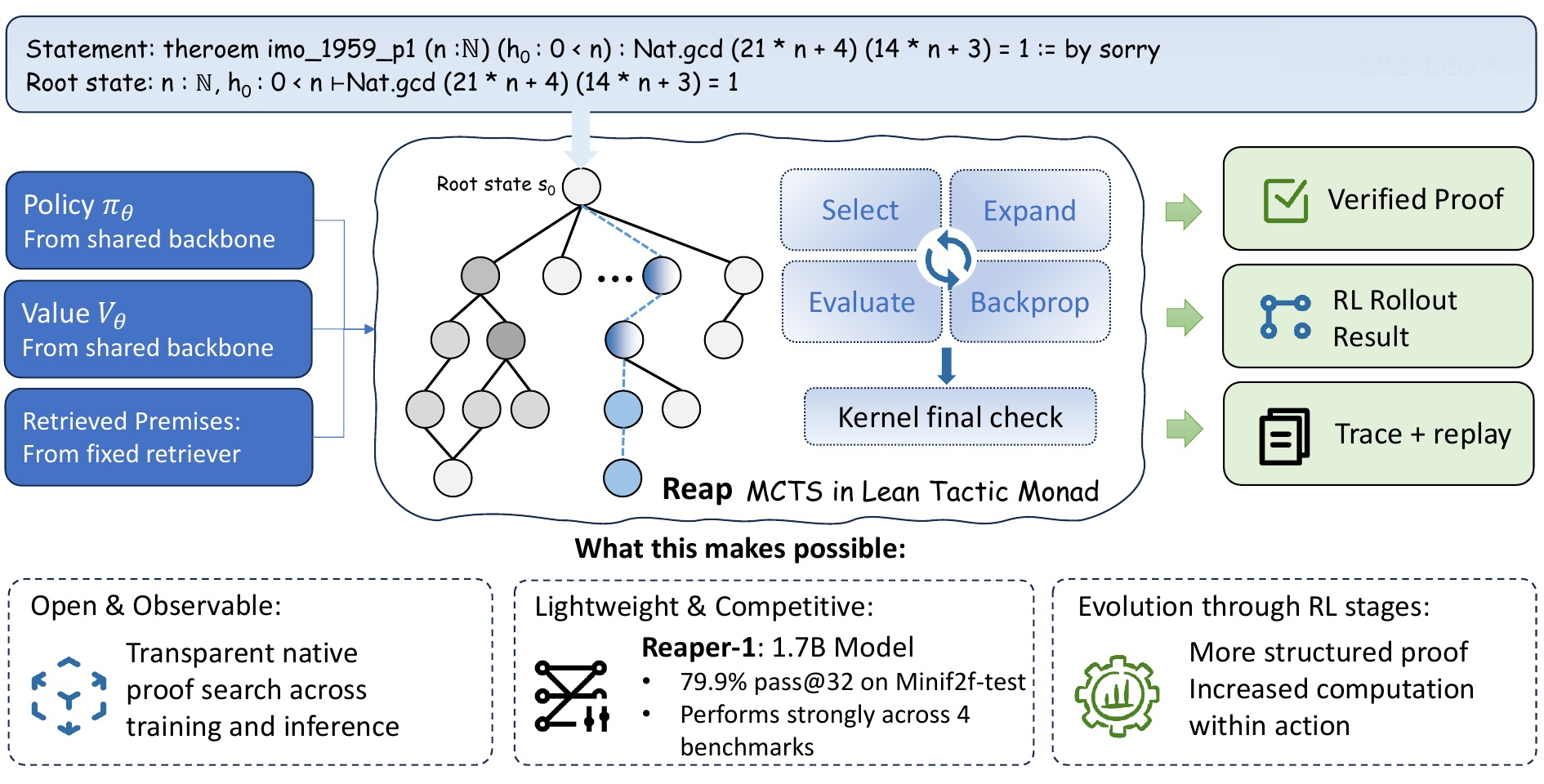}
\end{center}
\caption{Overview of Reap's proof-search and learning interface. A shared policy--value backbone and a fixed premise retriever provide guidance to Monte Carlo Tree Search running inside Lean's tactic monad, where native proof-state transitions are selected, expanded, evaluated, and backed up. Successful proofs are checked by Lean's kernel and exposed through traces and replay; the same interface supports reinforcement-learning rollouts and inference, enabling transparent analysis of how training changes proof structure and computation within each action.}
\label{fig:intro}
\end{figure}

Using this framework, we train \textbf{Reaper-1}, a lightweight 1.7B-parameter policy--value model through continued pre-training, supervised tactic learning, and iterative MCTS-guided reinforcement learning. With a separate premise retriever following the LeanSearch-PS recipe~\citep{realprover}, Reaper-1 achieves \textbf{79.9\%} pass@32 on miniF2F-test, establishing state-of-the-art performance among open provers of comparable size. This system altogether lowers the hardware barrier and compute budget to inference and experimentation.

Beyond benchmark performance, Reap allows us to study how reinforcement learning reshapes proof search. 
Our analysis reveals that reinforcement learning changes both the organization of proofs and the granularity of interaction with Lean. Later checkpoints produce shorter top-level proof trajectories while increasingly using structural tactics to introduce intermediate claims and organize local reasoning. Together with our qualitative case studies, this pattern suggests that the model becomes better at identifying useful intermediate structural lemmas and concentrating proof attempts on steps that make substantive progress, reducing reliance on repeated local manipulations. At the same time, individual actions become more compound, incorporating multiple proof operations and solver calls. Shorter trajectories therefore reflect a combination of more focused proof construction and greater computation within each action. These findings highlight the need to examine proof structure and execution cost together when interpreting the benefits of reinforcement learning for proof search.

Our main contributions are as follows:
\begin{itemize}
\item We introduce \textbf{Reap}, an open and observable MCTS proof-search framework implemented directly as a Lean~4 tactic, supporting native proof-state transitions, AND--OR search, proof reconstruction, and replay verification.
\item We develop the \textbf{Scythe-rl} training pipeline and use it to train \textbf{Reaper-1}, a lightweight 1.7B-parameter prover backbone. Integrated with Reap and a separate premise retriever, it achieves 79.9\% pass@32 on miniF2F-test and strongly transfer across 4 additional benchmarks, reaching state-of-the-art performance among models of comparable size.
\item Enabled by the high observability of our framework, we provide statistical analyses and detailed case studies to characterize how proof-search behavior evolves during MCTS-guided learning, revealing a transition toward search-interaction compression.
\end{itemize}

\section{Related Work}

\paragraph{LLM-based Formal Theorem Proving.}
Large language models have substantially advanced formal theorem proving in Lean. Recent systems generally follow either whole-proof generation or step-level interaction with the proof assistant. Representative whole-proof provers include DeepSeek-Prover-V2~\citep{deepseekproverv2}, Kimina-Prover~\citep{kiminaprover}, Goedel-Prover-V2~\citep{goedelproverv2}, and Pythagoras-Prover~\citep{pythagorasprover}. Step-level provers instead predict tactics conditioned on the current Lean proof state, as in HunyuanProver~\citep{hunyuanprover}, InternLM2.5-StepProver~\citep{internlm25stepprover}, REAL-Prover~\citep{realprover}, BFS-Prover-V2~\citep{bfsproverv2}, and OptProver~\citep{optprover}. Retrieval-augmented systems such as LeanDojo/ReProver~\citep{leandojo} and REAL-Prover retrieve relevant library declarations to guide proof generation. Reap follows the step-level paradigm and performs proof search over Lean states.

\paragraph{Search for Theorem Proving.}
Search is a key component of step-level theorem proving. Best-first search is widely used in recent neural provers~\citep{bfsproverv2}. Several works have also explored MCTS-style search. HyperTree Proof Search~\citep{htps} introduces an AlphaZero-inspired search procedure over the AND--OR structure of proof obligations. DeepSeek-Prover-V1.5~\citep{deepseekproverv15} proposes RMaxTS, an MCTS variant that encourages exploration over intermediate tactic states. AlphaProof~\citep{alphaproof} further combines policy--value-guided tree search with reinforcement learning at large scale. Reap builds on this line of work, but implements the MCTS rollout process directly inside Lean so that search operates on native proof states and can be invoked as a Lean tactic.

\paragraph{Reinforcement Learning for Formal Proving.}
Reinforcement learning has increasingly been used to improve formal provers through feedback from proof assistants. HTPS~\citep{htps} combines proof search with online learning from discovered proofs, while DeepSeek-Prover-V1.5~\citep{deepseekproverv15} introduces reinforcement learning from proof-assistant feedback. AlphaProof~\citep{alphaproof} couples policy--value-guided tree search with iterative reinforcement learning, which provides the primary methodological inspiration for Reap. Recent systems further explore interleaved thinking and proving, self-play, verifier-integrated reinforcement learning, and large-scale test-time search~\citep{leanstar,stp,leanabellv2,seedprover}.

\section{Methodology}

\subsection{Reap: Lean-Native Monte Carlo Tree Search}
\label{sec:reap-mcts}

Reap implements a policy--value-guided Monte Carlo Tree Search inspired by AlphaProof~\citep{alphaproof} and adapted to tactic-level theorem proving in Lean. Its key design is to run the search loop directly inside Lean's tactic monad. Proof-state transitions, tree construction, and verification remain inside Lean, allowing the same implementation to serve both as a training rollout engine and as an invocable Lean tactic. Neural policy and value evaluations are served externally: the policy proposes tactics, while the value estimates negative remaining proof-search cost and guides MCTS toward shorter successful continuations. Furthermore, Reap optionally records structured action-level logs covering model calls, tactic evaluation, and timing, enabling the analyses in Section~\ref{sec:rl-effect-study}. Implementation details of the search procedure are available in the public Reap repository at \url{https://github.com/frenzymath/reap}.

\begin{figure}[!htbp]
\centering
\includegraphics[width=\linewidth]{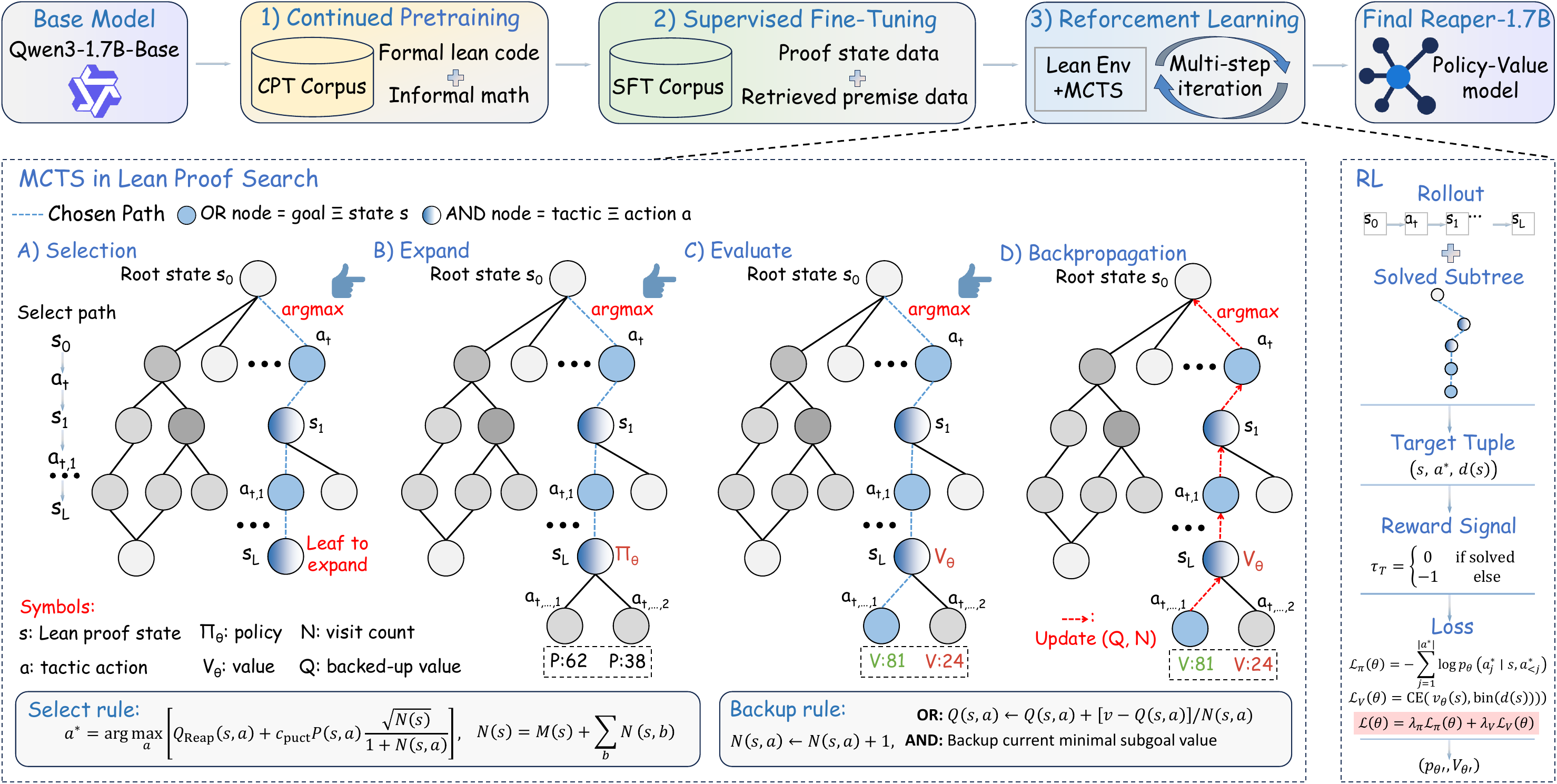}
\caption{Overview of Reap's training pipeline and Lean-native MCTS. In the backup rule, OR/AND labels refer to the successor node's type. Both $Q(s,a)$ and the return $v$ account for the edge cost: one for tactic edges and zero for synthetic focus edges. Here, $M(s)$ counts direct policy--value evaluations.}
\label{fig:method}
\end{figure}

\paragraph{Native proof states and tactic evaluation.}
The proof search is implemented natively in Lean through its metaprogramming
API. State transitions manipulate Lean's proof-state objects directly and use
the built-in save/restore primitives. Each node stores a
\texttt{Tactic.SavedState}, and each tactic edge represents a successful transition
induced by a Lean tactic. Before elaboration, Reap filters tactic syntax that
is unsuitable for a proof-state transition such as \texttt{sorry} and \texttt{simp?}. Tactic execution is
bounded by both Lean heartbeats and a wall-clock timeout. Only tactics that elaborate and
execute successfully are incorporated into the search tree, together with
their resulting proof states.

\paragraph{AND--OR proof search.}
Reap represents tactic choices with OR nodes and independent proof obligations with AND nodes. An OR node is considered solved once any child is solved, whereas an AND node requires that all its children be solved. Multiple goals are decomposed into an AND node only when they contain no unresolved expression metavariables; otherwise, they remain coupled in a single OR-state.

Node values estimate negative remaining proof-search cost. Each tactic edge incurs unit cost, while the synthetic focus edges connecting an AND node to its subgoals incur zero cost. OR nodes maintain visit-averaged value estimates. At an AND node, both the stored value and the value propagated to its parent are the minimum of the current values of its evaluated, unresolved children, with zero as the default for solved or unevaluated children. Thus, a fully solved AND node has value zero, and the estimated remaining cost of an AND node is determined by its hardest unresolved subgoal rather than the sum of its subgoal costs.

\paragraph{Progressive sampling and PUCT.}
Reap uses progressive sampling~\citep{alphaproof} to balance generating new tactics and revisiting
existing children. Let $N(s)$ be the visit count and $M(s)$ the number of
policy--value evaluations at state $s$. New tactics are sampled according to

\begin{equation}
M(s) \leq C\,N(s)^{\alpha}.
\label{eq:progressive-sampling}
\end{equation}

Meanwhile, Reap selects a child using the PUCT score~\citep{silver2018alphazero,alphaproof}

\begin{equation}
\operatorname{PUCT}(s,a)=\gamma^{-V(s,a)-1}+
c(N(s))P(s,a)\frac{\sqrt{N(s)}}{1+N(s,a)},
\label{eq:puct}
\end{equation}

where the exploration coefficient is

\begin{equation}
c(N)=c_{\mathrm{init}}+
\log\frac{N+c_{\mathrm{base}}+1}{c_{\mathrm{base}}},
\label{eq:puct-exploration}
\end{equation}

Here, $P(s,a)$ is the normalized policy prior, and
$V(s,a)=\bar v(s')-d(a)$ is the child node's current value estimate after
subtracting the edge cost, where $s'$ is the child state, $d(a)=1$ for tactic
edges, and $d(a)=0$ for synthetic focus edges. For visited children of OR
nodes, Eq.~(\ref{eq:puct}) favors lower estimated remaining costs. The
exploitation term is initialized to zero for unvisited children. AND nodes
instead use $1-\gamma^{-V(s,a)-1}$ for visited, unresolved children, retaining
the same exploration term and excluding solved children, to prioritize harder
subgoals; their unvisited children have exploitation term one.

\paragraph{State deduplication.}
When expanding an OR node, Reap executes each generated tactic and computes a serialized pretty-printed proof-state. If this key matches that of a state already on the current MCTS path, the candidate is discarded. If it matches an existing child of the current node, the candidate's prior weight is added to the existing edge and no new node is created. Every generated sample, including repeated tactic text within a sampling batch, contributes its prior weight when accepted. These weights are obtained by exponentiating temperature-scaled tactic log-probabilities and are normalized over the node's outgoing edges to obtain $P(s,a)$.

\paragraph{Proof reconstruction and final verification.}
A root node marked as solved by MCTS does not necessarily imply that the original theorem has been fully proved, as tactic execution may introduce auxiliary declarations or defer proof obligations. Reap therefore restores the original proof state, replays the selected tactics from the root, and independently validates the reconstructed proof using Lean’s ordinary declaration-checking pipeline to ensure the soundness of the generated proof.

\subsection{Reaper-1: Model and Training}

Reaper-1 is a 1.7B-parameter policy--value model initialized from Qwen3-1.7B-Base. Training proceeds in three main stages: continued pre-training (CPT) adapts the base model to Lean code and formal mathematics, supervised fine-tuning (SFT) learns tactic prediction from proof-state transitions, and iterative reinforcement learning improves both the policy and value estimates using trajectories generated by Reap. Training details are provided in the main text and Appendix~\ref{app:experimental-details}. Reaper-1 model weights are publicly available at \url{https://huggingface.co/FrenzyMath/Reaper-1}. The training pipeline code implemented in Scythe-rl is being prepared for public release.

\paragraph{Architecture.}
Reaper-1 uses a shared Transformer backbone for both policy and value prediction. The policy head is the standard language-modeling head and autoregressively generates Lean tactics conditioned on the current proof state and retrieved premises. An additional value head estimates the remaining proof-search cost of the current state using a discretized value representation. In the current implementation, policy and value predictions are requested in parallel.

\paragraph{Continued pre-training.}
Starting from Qwen3-1.7B-Base, we continue pre-training on Lean and informal mathematics data, combining FineLeanCorpus~\citep{criticlean}, LeanWorkbook~\citep{leanworkbook}, NuminaMath-Lean~\citep{kiminaprover}, DeepSeek-Prover-V1 data~\citep{deepseekproverv1}, and Mathlib v4.28.0-rc1. For Mathlib, we include source code together with term and declaration data extracted by jixia~\citep{jixia}, and additionally use internally annotated abstract-algebra data.

\paragraph{Supervised fine-tuning.}
We construct SFT data primarily from existing Lean proofs and Mathlib. jixia~\citep{jixia} extracts elaboration traces and proof-state transitions, which scythe converts into step-level examples containing tactics, premises, and value-related information. For each state, PS-2 retrieves relevant library declarations, and the model is trained to predict the next tactic conditioned on the state and retrieved premises. We further normalize structural tactics such as \texttt{have} and \texttt{suffices} into the Reap transition format (Appendix~\ref{app:tactic-normalization}). These state--premise--tactic examples initialize the policy before reinforcement learning.

\paragraph{Policy--value training objective.}
After each rollout stage, we collect successful trajectories that pass Lean replay and final verification and convert them into a replay dataset
$\mathcal{D}_{\mathrm{RL}}={(s,a,y)}$, where $a$ is the successful tactic at proof state $s$ and $y$ is its value target. The policy is trained autoregressively on the complete tactic sequence. For the value target, each valid tactic transition incurs unit search cost; the remaining proof cost is backed up through the corresponding AND--OR tree and discretized into $K=32$ bins. We jointly optimize
\begin{equation}
\begin{aligned}
\mathcal{L}
={}&
\lambda_P
\big(\underbrace{
-\mathbb{E}_{(s,a,y)\sim\mathcal{D}_{\mathrm{RL}}}
\left[\log \pi_\theta(a\mid s)\right]
}_{\mathcal{L}_{\mathrm{policy}}}\big)
\\
&+
\lambda_V
\big(\underbrace{
-\mathbb{E}_{(s,a,y)\sim\mathcal{D}_{\mathrm{RL}}}
\left[\log p_\theta(y\mid s)\right]
}_{\mathcal{L}_{\mathrm{value}}}\big).
\end{aligned}
\label{eq:joint-policy-value-loss}
\end{equation}
where $\pi_\theta(a\mid s)$ is the autoregressive probability of tactic $a$ and
$p_\theta(y\mid s)$ is the predicted probability of its discretized value target.

\paragraph{Iterative reinforcement learning.} 
Our RL stage follows an AlphaProof-style expert-iteration formulation: MCTS acts as a policy-improvement operator, and the network is trained on search-generated verified trajectories and backed-up value targets.
Starting from the SFT checkpoint, we perform five stages of MCTS-guided reinforcement learning on approximately 25K curated training problems extracted from AoPS. During RL1--RL4, each stage performs a full pass over the training corpus while progressively increasing the MCTS simulation budget, starting from 512 simulations in RL1. After performance on training corpus begins to saturate, RL5 switches to a batched multi-rollout regime with 8,192 simulations and pass@8. Beginning with B13, we additionally introduce group-wise value-based trajectory selection. For each theorem, we retain at most two successful trajectories with the smallest absolute backed-up root values. The complete RL procedure uses 60 NVIDIA A100 GPU days.

\paragraph{Premise selector.}
Following retrieval-augmented proving practice, we condition tactic generation
on relevant Mathlib declarations. Our default selector, PS-2, is trained
separately following the LeanSearch-PS recipe~\citep{realprover}, with an updated
Mathlib v4.28.0-rc1 snapshot and revised data filters. It remains fixed during
prover RL and supplies library declarations to Reaper-1, which performs tactic
generation and value prediction. Data construction, the selector-training
objective, and hyperparameters are detailed in Appendix~\ref{app:ps-data}.

\section{Experiments}
\label{sec:experiments}

\subsection{Experimental Setup}

\paragraph{Benchmarks.}
We evaluate Reaper-1 on five Lean theorem-proving benchmarks: \textbf{miniF2F-test}~\citep{minif2f}, our primary benchmark for competition-style mathematics; \textbf{ProofNet-test}~\citep{proofnet}; \textbf{FATE-M}~\citep{realprover} (150-problem snapshot, commit 4eb33c8); \textbf{PutnamBench}~\citep{putnambench} (660-problem snapshot, commit 6df1d4c); and \textbf{CombiBench}~\citep{combibench}. We decontaminate all training corpora against every evaluation benchmark using 10-gram filtering, and use these benchmarks only for retrospective evaluation, never for training or model selection.

\paragraph{Baselines and evaluation.}
We compare with representative whole-proof and step-level Lean provers. Since published systems use different inference budgets, Table~\ref{tab:main_results} reports each result with its original evaluation budget rather than treating the comparisons as compute-matched. Reaper-1 is evaluated using the final RL policy--value checkpoint and the default premise-selection pipeline. Unless otherwise specified, controlled miniF2F experiments use 8,192 MCTS simulations per search and a maximum policy-generation length of 1,024 tokens. Under this configuration, evaluating all 244 miniF2F-test problems at pass@32 only takes approximately 3 hours of wall-clock time using 32 NVIDIA A100 GPUs and 800 CPU cores. The total inference time of the premise selector is approximately 8.6\% of the total inference time in the experiment.
All reported successes are replayed and verified by Lean.

\paragraph{Premise selection.}
Premise retrieval is provided by a separate, modular component rather than the Reaper-1 network. Our default setup uses PS-2, an independently trained Qwen3-Embedding-8B selector, which is neither jointly optimized with Reaper-1 nor adapted during RL. We evaluate inference-time replacement with Lean's rule-based selectors, without retraining the policy--value model, in Table~\ref{tab:retriever-replacement}.

\subsection{Main Results}

Despite using a 1.7B-parameter generative policy--value model, Reaper-1 is highly competitive with much
larger formal provers. On miniF2F-test, it reaches 79.9\% pass@32, improving over
the 1.7B Kimina-Prover-RL baseline by 3.3 percentage points and outperforming
several 7B step-level and whole-proof systems. The advantage is also evident beyond the primary benchmark: Reaper-1 obtains 33.9\% on ProofNet-test, compared with 23.7\% for REAL-Prover, and reaches 81.3\% on
FATE-M. On the harder suites, it solves 21/660 PutnamBench problems (pass@32) and 7/100
CombiBench problems (pass@16). Across checkpoints accumulated during our
iterative training procedure, the union of solved miniF2F-test problems reaches
81.1\%.

\begin{table}[!htbp]
\caption{Main benchmark results. ``--'' indicates that the corresponding result was not reported by the original work.}
\label{tab:main_results}
\centering
\small
\resizebox{\linewidth}{!}{
\begin{tabular}{lcccccc}
\toprule
\textbf{System} & \textbf{Budget} & \textbf{miniF2F} & \textbf{ProofNet} &
\textbf{FATE-M} & \textbf{PutnamBench} & \textbf{CombiBench} \\
\midrule
\multicolumn{7}{l}{\textit{Whole-proof systems}} \\
DeepSeek-Prover-V2-7B~\citep{deepseekproverv2}      & pass@32 & 75.6\% & --    & --    & 9/660  & 6 (pass@16) \\
Goedel-Prover-V2-8B~\citep{goedelproverv2}          & pass@32 & 84.6\% & --    & --    & 25/660 & -- \\
Kimina-Prover-7B~\citep{kiminaprover}               & pass@32 & 78.3\% & --    & --    & --     & 7 (pass@16) \\
Kimina-Prover-RL-1.7B~\citep{kiminaprover}     & pass@32 & 76.6\% & --    & --    & --     & -- \\
Pythagoras-Prover-4B~\citep{pythagorasprover}       & pass@32 & 86.1\% & --    & --    & --     & -- \\
\midrule
\multicolumn{7}{l}{\textit{Step-level systems}} \\
HunyuanProver~\citep{hunyuanprover}                 & pass@8  & 68.4\% & --    & --    & --     & -- \\
ABEL                          & pass@32 & 41.3\% & --    & --    & 8/660  & -- \\
REAL-Prover~\citep{realprover}                      & pass@64 & 54.1\% & 23.7\%& 56.7\%& --     & -- \\
InternLM2.5-StepProver-7B~\citep{internlm25stepprover} & pass@32 & 63.8\% & 23.6\%& -- & -- & -- \\
BFS-Prover-V2-7B~\citep{bfsproverv2}                & Accumulative & 82.4\% & --    & --    & --     & -- \\
\midrule
\textbf{Reaper-1 (1.7B)}      & pass@32 & \textbf{79.9\%} & \textbf{33.9\%} &
\textbf{81.3\%} & \textbf{21/660} & \textbf{7/100 (pass@16)} \\
\bottomrule
\end{tabular}}
\end{table}

\subsection{Reinforcement-Learning Progression}

\begin{figure}[!htbp]
\centering
\includegraphics[width=\linewidth]{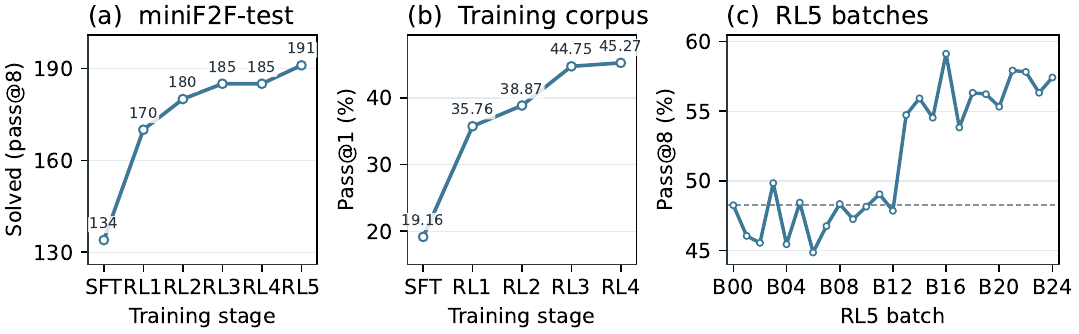}
\caption{Reinforcement-learning progression. (a) Retrospective evaluation on miniF2F-test at
8,192 MCTS simulations and pass@8. (b) Pass@1 on the full RL training
corpus. (c) Sequential RL5 small-batch evaluation; the dashed horizontal line
marks the B00 value (48.26\%).}
\label{fig:rl_progression}
\end{figure}

Figure~\ref{fig:rl_progression} retrospectively tracks the effect of iterative reinforcement learning from the SFT checkpoint. On miniF2F-test, the number of problems solved under pass@8 increases from 134 for SFT to 191 after RL5, with the largest early improvement occurring during the first RL stage; a similar trend is observed on the full RL training corpus, where pass@1 rises from 19.16\% for SFT to 45.27\% after RL4. As performance begins to plateau during RL3--RL4, RL5 switches from full-corpus iterations to sequential small-batch updates with a revised rollout and trajectory-selection regime. Despite substantial batch-to-batch variation, performance improves again in the later RL5 batches, indicating that the sequential training regime is associated with extending prover capability after the earlier full-corpus iterations begin to saturate.

\subsection{Ablation Studies}
\label{sec:ablation}

\paragraph{Value head.}
We isolate the effect of the value head guidance during proof search. In the ablated configuration, the neural
value prediction is disabled by fixing the predicted value of every state to
zero, while the rest of the search pipeline remains unchanged. As shown in
Table~\ref{tab:value_ablation}, on the 193 problems solved by both RL5 and RL5 without the value head, RL5 produces shorter successful proofs in the sense of both average tactic steps and an average root value closer to zero. This suggests that enabling value guidance biases MCTS toward shorter successful  trajectories.

\begin{table}[!htbp]
\caption{Value-head ablation on miniF2F-test with 8,192 simulations and
pass@32. Tactic steps are computed by first averaging proved passes within each
theorem and then averaging equally across theorems. Average tactic steps and root value are computed on the both-solved 193 problems.}
\label{tab:value_ablation}
\centering
\small
\begin{tabular}{lccc}
\toprule
\textbf{Model} & \textbf{Solved} & \textbf{Avg. tactic steps} &
\textbf{Avg. root value} \\
\midrule
RL5                    & 195 & 16.4553 & -9.9174 \\
RL5 w/o value head& 193 & 17.4435          & -11.0221 \\
\bottomrule
\end{tabular}
\end{table}

\FloatBarrier

\begin{table}[!htbp]
\caption{Inference-time retriever replacement on miniF2F-test with the same
RL5 checkpoint and 8,192 MCTS simulations per search. PS-2 is neural;
MePo and SineQuoNon are rule-based. Parentheses give solved/total problems.}
\label{tab:retriever-replacement}
\centering
\small
\begin{tabular}{lccc}
\toprule
\textbf{Retriever} & \textbf{PS-2} & \textbf{MePo} & \textbf{SineQuoNon} \\
\midrule
\textbf{Pass@32} & 79.9\% (195/244) & 67.6\% (165/244) & 75.8\% (185/244) \\
\bottomrule
\end{tabular}
\end{table}

\FloatBarrier

\paragraph{Replacing the premise retriever.}
We replace PS-2 at inference with Lean's built-in rule-based selectors,
MePo~\citep{mepo} and SineQuoNon~\citep{sine}, keeping the final RL5
policy--value checkpoint and search configuration fixed.
Despite being trained with PS-2 premises, Reaper-1 achieves 75.8\% pass@32
with SineQuoNon without additional training
(Table~\ref{tab:retriever-replacement}). This is 10 problems
(4.1 percentage points) below the 79.9\% obtained with PS-2; MePo achieves
67.6\%. These results demonstrate that the learned prover can operate
effectively with a rule-based retriever on this benchmark, supporting
inference-time replacement of the neural retrieval component.

\subsection{Search Scaling}

We next examine how Reaper-1 scales with the MCTS simulation budget. Table~\ref{tab:search-scaling} reports miniF2F-test performance under pass@32 while varying the maximum number of MCTS simulations allowed for each independent search. Reaper-1 already solves 188 of 244 problems (77.05\%) with only 512 simulations. Coverage then begins to saturate: both 2,048 and 4,096 simulations solve the same 194 problems, while the full 8,192-simulation budget adds only one additional theorem.
The scaling curve therefore exhibits clear diminishing returns in theorem coverage. The strong performance at relatively small budgets suggests that Reaper-1's final result is not solely a consequence of the largest test-time search budget.

\begin{table}[!htbp]
\caption{Search scaling of Reaper-1 on miniF2F-test under pass@32.}
\label{tab:search-scaling}
\centering
\small
\begin{tabular}{rccccc}
\toprule
\textbf{MCTS budget} & 512 & 1024 & 2048 & 4096 & 8192 \\
\midrule
\textbf{Pass@32} & 77.05\% & 78.28\% & 79.51\% & 79.51\% & 79.92\% \\
\bottomrule
\end{tabular}
\end{table}

\FloatBarrier

\subsection{How Reinforcement Learning Reshapes Proof Search}
\label{sec:rl-effect-study}

Beyond improving theorem coverage, reinforcement learning also changes how the
prover interacts with search. We compare the SFT, RL4, and RL5 checkpoints under
the same miniF2F setting. To control for theorem difficulty, proof-level
statistics are computed on the 151 problems solved by all three checkpoints.
Full measurement protocols, statistical analyses, and qualitative examples are
deferred to Appendices~\ref{app:rl-effect-study} and
\ref{app:case-proof-evolution}.

\begin{table}[!htbp]
\caption{
How reinforcement learning reshapes proof search.
Coverage uses all 244 miniF2F-test problems; proof and action statistics use the
151-problem common solved set.
}
\label{tab:rl-effect-summary}
\centering
\small
\begin{tabular}{lrrr}
\toprule
\textbf{Metric} & \textbf{SFT} & \textbf{RL4} & \textbf{RL5} \\
\midrule
Pass@32 & 62.3\% & 78.3\% & 79.9\% \\
Mean proof steps & 21.7 & 11.2 & 5.2 \\
Mean search-tree nodes & 2,368 & 162 & 119 \\
\midrule
Median action characters & 19 & 23 & 66 \\
Tactic identifiers/action & 1.38 & 1.38 & 2.80 \\
Actions containing \texttt{<;>} & 2.9\% & 4.5\% & 23.9\% \\
Median evaluation latency (ms) & 12.0 & 15.2 & 43.2 \\
\bottomrule
\end{tabular}
\end{table}

\FloatBarrier

As shown in Table~\ref{tab:rl-effect-summary}, RL substantially improves
coverage while compressing successful search into shorter trajectories and much
smaller trees. The later checkpoints also exhibit a clear change in proof
organization: actions become longer and structurally denser, with greater use
of compound tactic programs. Together with the qualitative cases in
Appendix~\ref{app:case-proof-evolution}, this suggests that the policy
increasingly organizes proofs around useful intermediate structure and directs
search toward steps that make substantive progress, rather than relying on
repeated local manipulations.

This compression does not imply less underlying computation. As actions become
more compound, their Lean evaluation cost also increases. We therefore describe
the observed transition as \emph{search-interaction compression}: RL reduces the
number of top-level search decisions while shifting more proof construction and
solver work into each individual action.

\section{Limitations and Future Work}
\label{sec:limitations}

Reaper-1 uses the number of tactic-level search actions as a proxy for proof cost, which does not always reflect the actual complexity of a Lean proof step. During later RL stages, we observe that the model increasingly exploits Lean constructs that package multiple proof operations into a single search action. Such constructions can beneficially compress local reasoning and reduce search depth, but may also lower the apparent trajectory cost without reducing the underlying computation; sufficiently complex actions can increase Lean execution latency and reduce MCTS exploration efficiency. This limitation suggests that future objectives could incorporate tactic execution time, generated-token length, or finer-grained measures of action complexity. Finally, our experiments focus on Lean~4 and a single 1.7B policy--value model, so the extent to which these search and training dynamics generalize to substantially larger models or other proof assistants remains to be studied.

\section{Conclusion}

We introduced \textbf{Reap}, an open Monte Carlo Tree Search framework for formal theorem proving whose semantics-critical search loop runs directly inside Lean, together with \textbf{Reaper-1}, a 1.7B-parameter policy--value model trained through continued pre-training, supervised tactic learning, and iterative MCTS-guided reinforcement learning. Despite its modest scale, Reaper-1 with a separate
premise retriever achieves strong performance across multiple Lean benchmarks, including 79.9\% pass@32 on miniF2F-test, while remaining competitive with substantially larger provers. Beyond benchmark coverage, Reap provides a lightweight and observable platform for studying how neural policies, value estimation, formal verification, and tree search interact during theorem proving. Our analysis suggests that search-based RL can refine proof strategy, enabling the prover to make effective use of auxiliary lemmas and pursue more direct routes to a complete argument. As individual tactics take on a larger role in carrying out local reasoning, these advances also call for closer attention to their computational demands. By making such learning dynamics accessible to systematic study, Reap supports further research on training objectives that encourage meaningful proof progress.

\section*{Acknowledgments}
This work is supported in part by the National Key R\&D Program of China grant 2024YFA1014000, the Fundamental and Interdisciplinary Disciplines Breakthrough Plan of the Ministry of Education of China (JYB2025XDXM113), and the New Cornerstone Investigator Program.

\subsection*{AI use statement}

In this work, generative AI tools were used to assist with manuscript writing and organization based on detailed content provided by the authors, literature search, implementation of algorithmic components, and statistical analysis for the RL study. All AI-assisted outputs were reviewed and validated by the authors against the relevant source materials, code, and experimental data. The authors take full responsibility for the methodology, reported results, analyses, and final content of the paper.

\bibliography{references}

@article{hunyuanprover,
  title   = {{HunyuanProver}: A Scalable Data Synthesis Framework and Guided Tree Search for Automated Theorem Proving},
  author  = {Li, Yang and Du, Dong and Song, Linfeng and Li, Chen and Wang, Weikang and Yang, Tao and Mi, Haitao},
  journal = {arXiv preprint arXiv:2412.20735},
  year    = {2024}
}

@inproceedings{internlm25stepprover,
  title     = {{InternLM2.5-StepProver}: Advancing Automated Theorem Proving via Critic-Guided Search},
  author    = {Wu, Zijian and Huang, Suozhi and Zhou, Zhejian and Ying, Huaiyuan and Yuan, Zheng and Zhang, Wenwei and Lin, Dahua and Chen, Kai},
  booktitle = {2nd AI for Math Workshop @ ICML 2025},
  year      = {2025},
  url       = {https://openreview.net/forum?id=qwCqeIg5iI}
}

@inproceedings{deepseekproverv15,
  title   = {{DeepSeek-Prover-V1.5}: Harnessing Proof Assistant Feedback for Reinforcement Learning and Monte-Carlo Tree Search},
  author  = {Xin, Huajian and Ren, Z. Z. and Song, Junxiao and Shao, Zhihong and Zhao, Wanjia and Wang, Haocheng and Liu, Bo and Zhang, Liyue and Lu, Xuan and Du, Qiushi and Gao, Wenjun and Zhang, Haowei and Zhu, Qihao and Yang, Dejian and Gou, Zhibin and Wu, Z. F. and Luo, Fuli and Ruan, Chong},
  booktitle = {The Thirteenth International Conference on Learning Representations},
  year    = {2025},
  url     = {https://openreview.net/forum?id=I4YAIwrsXa}
}

@article{deepseekproverv2,
  title   = {{DeepSeek-Prover-V2}: Advancing Formal Mathematical Reasoning via Reinforcement Learning for Subgoal Decomposition},
  author  = {Ren, Z. Z. and Shao, Zhihong and Song, Junxiao and Xin, Huajian and Wang, Haocheng and Zhao, Wanjia and Zhang, Liyue and Fu, Zhe and Zhu, Qihao and Yang, Dejian and Wu, Z. F. and Gou, Zhibin and Ma, Shirong and Tang, Hongxuan and Liu, Yuxuan and Gao, Wenjun and Guo, Daya and Ruan, Chong},
  journal = {arXiv preprint arXiv:2504.21801},
  year    = {2025}
}

@article{kiminaprover,
  title   = {{Kimina-Prover} Preview: Towards Large Formal Reasoning Models with Reinforcement Learning},
  author  = {Wang, Haiming and Unsal, Mert and Lin, Xiaohan and Baksys, Mantas and Liu, Junqi and Dos Santos, Marco and Sung, Flood and Vinyes, Marina and Ying, Zhenzhe and Zhu, Zekai and Lu, Jianqiao and de Saxc{\'e}, Hugues and Bailey, Bolton and Song, Chendong and Xiao, Chenjun and Zhang, Dehao and Zhang, Ebony and Pu, Frederick and Zhu, Han and Liu, Jiawei and Bayer, Jonas and Michel, Julien and Yu, Longhui and Dreyfus-Schmidt, L{\'e}o and Tunstall, Lewis and Pagani, Luigi and Machado, Moreira and Bourigault, Pauline and Wang, Ran and Polu, Stanislas and Barroyer, Thibaut and Li, Wen-Ding and Niu, Yazhe and Fleureau, Yann and Hu, Yangyang and Yu, Zhouliang and Wang, Zihan and Yang, Zhilin and Liu, Zhengying and Li, Jia},
  journal = {arXiv preprint arXiv:2504.11354},
  year    = {2025}
}

@inproceedings{goedelproverv2,
  title     = {{Goedel-Prover-V2}: Scaling Formal Theorem Proving with Scaffolded Data Synthesis and Self-Correction},
  author    = {Lin, Yong and Tang, Shange and Lyu, Bohan and Yang, Ziran and Chung, Jui-Hui and Zhao, Haoyu and Jiang, Lai and Geng, Yihan and Ge, Jiawei and Sun, Jingruo and Wu, Jiayun and Gesi, Jiri and Lu, Ximing and Acuna, David and Yang, Kaiyu and Lin, Hongzhou and Choi, Yejin and Chen, Danqi and Arora, Sanjeev and Jin, Chi},
  booktitle = {The Fourteenth International Conference on Learning Representations},
  year      = {2026},
  url       = {https://openreview.net/forum?id=j4C0nALrgK}
}

@article{realprover,
  title   = {{REAL-Prover}: Retrieval Augmented Lean Prover for Mathematical Reasoning},
  author  = {Shen, Ziju and Huang, Naohao and Yang, Fanyi and Wang, Yutong and Gao, Guoxiong and Xu, Tianyi and Jiang, Jiedong and He, Wanyi and Yang, Pu and Sun, Mengzhou and Ju, Haocheng and Wu, Peihao and Dai, Bryan and Dong, Bin},
  journal = {arXiv preprint arXiv:2505.20613},
  year    = {2025}
}

@article{pythagorasprover,
  title   = {{Pythagoras-Prover}: Advancing Efficient Formal Proving via Augmented Lean Formalisation},
  author  = {Leang, Joshua Ong Jun and Zhao, Zheng and Stoian, Mihaela C{\u{a}}t{\u{a}}lina and Xu, Qiyuan and Li, Haonan and Li, Wenda and Cohen, Shay B. and Giunchiglia, Eleonora},
  journal = {arXiv preprint arXiv:2606.12594},
  year    = {2026}
}

@inproceedings{optprover,
  title     = {{OptProver}: Bridging Olympiad and Optimization through Continual Training in Formal Theorem Proving},
  author    = {Li, Chenyi and Nie, Yanchen and Ming, Zhenyu and Zhang, Gong and Yuan, Kun and Wen, Zaiwen},
  booktitle = {Forty-third International Conference on Machine Learning},
  year      = {2026},
  url       = {https://openreview.net/forum?id=mkHp4ZW01l}
}

@inproceedings{bfsproverv2,
  title     = {{BFS-Prover-V2}: Scaling up Multi-Turn Off-Policy {RL} and Multi-Agent Tree Search for {LLM} Step-Provers},
  author    = {Xin, Ran and Zheng, Zeyu and Nie, Yanchen and Yuan, Kun and Xiao, Xia},
  booktitle = {Forty-third International Conference on Machine Learning},
  year      = {2026},
  url       = {https://openreview.net/forum?id=GPrlMiGg2w}
}

@inproceedings{leandojo,
  title     = {{LeanDojo}: Theorem Proving with Retrieval-Augmented Language Models},
  author    = {Yang, Kaiyu and Swope, Aidan M. and Gu, Alex and Chalamala, Rahul and Song, Peiyang and Yu, Shixing and Godil, Saad and Prenger, Ryan J. and Anandkumar, Animashree},
  booktitle = {Advances in Neural Information Processing Systems},
  volume    = {36},
  pages     = {21573--21612},
  year      = {2023},
  doi       = {10.52202/075280-0944},
  url       = {https://proceedings.neurips.cc/paper_files/paper/2023/hash/4441469427094f8873d0fecb0c4e1cee-Abstract-Datasets_and_Benchmarks.html}
}

@article{polu2020gptf,
  title   = {Generative Language Modeling for Automated Theorem Proving},
  author  = {Polu, Stanislas and Sutskever, Ilya},
  journal = {arXiv preprint arXiv:2009.03393},
  year    = {2020}
}

@inproceedings{htps,
  title     = {HyperTree Proof Search for Neural Theorem Proving},
  author    = {Guillaume Lample and Timothee Lacroix and Marie-Anne Lachaux and Aurelien Rodriguez and Amaury Hayat and Thibaut Lavril and Gabriel Ebner and Xavier Martinet},
  booktitle = {Advances in Neural Information Processing Systems},
  editor    = {Koyejo, S. and Mohamed, S. and Agarwal, A. and Belgrave, D. and Cho, K. and Oh, A.},
  volume    = {35},
  pages     = {26337--26349},
  year      = {2022},
  doi       = {10.52202/068431-1910},
  url       = {https://proceedings.neurips.cc/paper_files/paper/2022/hash/a8901c5e85fb8e1823bbf0f755053672-Abstract-Conference.html}
}

@article{alphaproof,
  title     = {Olympiad-level formal mathematical reasoning with reinforcement learning},
  author    = {Hubert, Thomas and Mehta, Rishi and Sartran, Laurent and Horv{\'a}th, Mikl{\'o}s Z. and {\v{Z}}u{\v{z}}i{\'c}, Goran and Wieser, Eric and Huang, Aja and Schrittwieser, Julian and Schroecker, Yannick and Masoom, Hussain and Bertolli, Ottavia and Zahavy, Tom and Mandhane, Amol and Yung, Jessica and Beloshapka, Iuliya and Ibarz, Borja and Veeriah, Vivek and Yu, Lei and Nash, Oliver and Lezeau, Paul and Mercuri, Salvatore and S{\"o}nne, Calle and Mehta, Bhavik and Davies, Alex and Zheng, Daniel and Pedregosa, Fabian and Li, Yin and von Glehn, Ingrid and Rowland, Mark and Albanie, Samuel and Velingker, Ameya and Schmitt, Simon and Lockhart, Edward and Hughes, Edward and Michalewski, Henryk and Sonnerat, Nicolas and Hassabis, Demis and Kohli, Pushmeet and Silver, David},
  journal   = {Nature},
  volume    = {651},
  number    = {8106},
  pages     = {607--613},
  year      = {2026},
  publisher = {Nature Publishing Group},
  doi       = {10.1038/s41586-025-09833-y}
}

@article{silver2018alphazero,
  title     = {A general reinforcement learning algorithm that masters chess, shogi, and {Go} through self-play},
  author    = {Silver, David and Hubert, Thomas and Schrittwieser, Julian and Antonoglou, Ioannis and Lai, Matthew and Guez, Arthur and Lanctot, Marc and Sifre, Laurent and Kumaran, Dharshan and Graepel, Thore and Lillicrap, Timothy and Simonyan, Karen and Hassabis, Demis},
  journal   = {Science},
  volume    = {362},
  number    = {6419},
  pages     = {1140--1144},
  year      = {2018},
  publisher = {American Association for the Advancement of Science},
  doi       = {10.1126/science.aar6404}
}

@inproceedings{leanstar,
  title     = {{Lean-STaR}: Learning to Interleave Thinking and Proving},
  author    = {Lin, Haohan and Sun, Zhiqing and Welleck, Sean and Yang, Yiming},
  booktitle = {The Thirteenth International Conference on Learning Representations},
  year      = {2025},
  url       = {https://openreview.net/forum?id=SOWZ59UyNc}
}

@inproceedings{stp,
  title     = {{STP}: Self-play {LLM} Theorem Provers with Iterative Conjecturing and Proving},
  author    = {Dong, Kefan and Ma, Tengyu},
  booktitle = {Proceedings of the 42nd International Conference on Machine Learning},
  series    = {Proceedings of Machine Learning Research},
  volume    = {267},
  pages     = {14114--14136},
  year      = {2025},
  publisher = {PMLR},
  url       = {https://proceedings.mlr.press/v267/dong25h.html}
}

@article{leanabellv2,
  title   = {{Leanabell-Prover-V2}: Verifier-integrated Reasoning for Formal Theorem Proving via Reinforcement Learning},
  author  = {Ji, Xingguang and Liu, Yahui and Wang, Qi and Zhang, Jingyuan and Yue, Yang and Shi, Rui and Sun, Chenxi and Zhang, Fuzheng and Zhou, Guorui and Gai, Kun},
  journal = {arXiv preprint arXiv:2507.08649},
  year    = {2025}
}

@article{seedprover,
  title   = {{Seed-Prover}: Deep and Broad Reasoning for Automated Theorem Proving},
  author  = {Chen, Luoxin and Gu, Jinming and Huang, Liankai and Huang, Wenhao and Jiang, Zhicheng and Jie, Allan and Jin, Xiaoran and Jin, Xing and Li, Chenggang and Ma, Kaijing and Ren, Cheng and Shen, Jiawei and Shi, Wenlei and Sun, Tong and Sun, He and Wang, Jiahui and Wang, Siran and Wang, Zhihong and Wei, Chenrui and Wei, Shufa and Wu, Yonghui and Wu, Yuchen and Xia, Yihang and Xin, Huajian and Yang, Fan and Ying, Huaiyuan and Yuan, Hongyi and Yuan, Zheng and Zhan, Tianyang and Zhang, Chi and Zhang, Yue and Zhang, Ge and Zhao, Tianyun and Zhao, Jianqiu and Zhou, Yichi and Zhu, Thomas Hanwen},
  journal = {arXiv preprint arXiv:2507.23726},
  year    = {2025}
}

@article{infonce,
  title   = {Representation Learning with Contrastive Predictive Coding},
  author  = {van den Oord, Aaron and Li, Yazhe and Vinyals, Oriol},
  journal = {arXiv preprint arXiv:1807.03748},
  year    = {2018}
}

@inproceedings{leanworkbook,
  title     = {{Lean Workbook}: A Large-Scale Lean Problem Set Formalized from Natural Language Math Problems},
  author    = {Ying, Huaiyuan and Wu, Zijian and Geng, Yihan and Wang, Jiayu and Lin, Dahua and Chen, Kai},
  booktitle = {Advances in Neural Information Processing Systems},
  volume    = {37},
  pages     = {105848--105863},
  year      = {2024},
  doi       = {10.52202/079017-3357}
}

@inproceedings{criticlean,
  title     = {{CriticLean}: Critic-Guided Reinforcement Learning for Mathematical Formalization},
  author    = {Peng, Zhongyuan and Yao, Yifan and Ma, Kaijing and Guo, Shuyue and Li, Yizhe and Zhang, Yichi and Zhang, Chenchen and Zhang, Yifan and Yu, Zhouliang and Li, Luming and Liu, Minghao and Xia, Yihang and Shen, Jiawei and Wu, Yuchen and Cao, Yixin and Zhang, Zhaoxiang and Huang, Wenhao and Liu, Jiaheng and Zhang, Ge},
  booktitle = {Proceedings of the 64th Annual Meeting of the Association for Computational Linguistics (Volume 1: Long Papers)},
  pages     = {3049--3088},
  year      = {2026},
  publisher = {Association for Computational Linguistics},
  doi       = {10.18653/v1/2026.acl-long.139},
  url       = {https://aclanthology.org/2026.acl-long.139/}
}

@article{deepseekproverv1,
  title   = {{DeepSeek-Prover}: Advancing Theorem Proving in {LLMs} through Large-Scale Synthetic Data},
  author  = {Xin, Huajian and Guo, Daya and Shao, Zhihong and Ren, Zhizhou and Zhu, Qihao and Liu, Bo and Ruan, Chong and Li, Wenda and Liang, Xiaodan},
  journal = {arXiv preprint arXiv:2405.14333},
  year    = {2024}
}

@inproceedings{minif2f,
  title     = {{MiniF2F}: A Cross-System Benchmark for Formal Olympiad-Level Mathematics},
  author    = {Zheng, Kunhao and Han, Jesse Michael and Polu, Stanislas},
  booktitle = {The Tenth International Conference on Learning Representations},
  year      = {2022},
  url       = {https://openreview.net/forum?id=9ZPegFuFTFv}
}

@article{proofnet,
  title   = {{ProofNet}: Autoformalizing and Formally Proving Undergraduate-Level Mathematics},
  author  = {Azerbayev, Zhangir and Piotrowski, Bartosz and Schoelkopf, Hailey and Ayers, Edward W. and Radev, Dragomir and Avigad, Jeremy},
  journal = {arXiv preprint arXiv:2302.12433},
  year    = {2023}
}

@inproceedings{putnambench,
  title     = {{PutnamBench}: Evaluating Neural Theorem-Provers on the Putnam Mathematical Competition},
  author    = {Tsoukalas, George and Lee, Jasper and Jennings, John and Xin, Jimmy and Ding, Michelle and Jennings, Michael and Thakur, Amitayush and Chaudhuri, Swarat},
  booktitle = {Advances in Neural Information Processing Systems},
  volume    = {37},
  pages     = {11545--11569},
  year      = {2024},
  doi       = {10.52202/079017-0368}
}

@inproceedings{combibench,
  title     = {{CombiBench}: Benchmarking {LLM} Capability for Combinatorial Mathematics},
  author    = {Liu, Junqi and Lin, Xiaohan and Bailey, Bolton and Bayer, Jonas and Dillies, Ya{\"e}l and Jiang, Weijie and Liang, Xiaodan and Soletskyi, Roman and Wang, Haiming and Xie, Yunzhou and Xiong, Beibei and Yang, Zhengfeng and Zhang, Jujian and Zhi, Lihong and Zhu, Zekai and Li, Jia and Liu, Zhengying},
  booktitle = {3rd AI for Math Workshop: Toward Self-Evolving Scientific Agents},
  year      = {2026},
  url       = {https://openreview.net/forum?id=qeZStJVYnc}
}

@article{mepo,
  title     = {Lightweight Relevance Filtering for Machine-Generated Resolution Problems},
  author    = {Meng, Jia and Paulson, Lawrence C.},
  journal   = {Journal of Applied Logic},
  volume    = {7},
  number    = {1},
  pages     = {41--57},
  year      = {2009},
  doi       = {10.1016/j.jal.2007.07.004}
}

@inproceedings{sine,
  title     = {Sine Qua Non for Large Theory Reasoning},
  author    = {Hoder, Kry{\v{s}}tof and Voronkov, Andrei},
  booktitle = {Automated Deduction -- {CADE-23}},
  series    = {Lecture Notes in Computer Science},
  volume    = {6803},
  pages     = {299--314},
  year      = {2011},
  publisher = {Springer},
  doi       = {10.1007/978-3-642-22438-6_23}
}

@misc{jixia,
  author       = {{FrenzyMath}},
  title        = {{Jixia}},
  howpublished = {GitHub repository},
  year         = {2026},
  url          = {https://github.com/frenzymath/jixia},
  note         = {Version v4.28.0-rc1},
  urldate      = {2026-09-07}
}
\bibliographystyle{references}

\appendix

\section{Experimental Details}
\label{app:experimental-details}

This section provides additional hyperparameter details for the training and
evaluation experiments in Section~\ref{sec:experiments}. Unless otherwise
specified, proof search uses the MCTS configuration described in Section~\ref{sec:reap-mcts}
with both policy and value prediction enabled and neural premise selection.
All experiments use Lean~v4.28.0-rc1 with the corresponding Mathlib snapshot,
and a proof is counted as successful only after final replay verification in
Lean.

Table~\ref{tab:search-hyperparameters} summarizes the search and inference
configuration used for the controlled miniF2F experiments with 8,192 MCTS
simulations. At each policy evaluation, Reaper-1 samples six candidate tactics
with temperature $0.99$ and a maximum generation length of 1,024 tokens, while
the premise selector retrieves eight declarations for the current proof state.

\begin{table}[!htbp]
\caption{MCTS and inference hyperparameters used in the controlled miniF2F
experiments.}
\label{tab:search-hyperparameters}
\centering
\small
\begin{tabular}{lc}
\toprule
\textbf{Parameter} & \textbf{Value} \\
\midrule
Search algorithm & MCTS \\
Maximum MCTS simulations & 8,192 \\
Maximum tree nodes & 65,536 \\
Independent attempts & 32 \\
Policy samples/state & 6 \\
Retrieved premises/state & 8 \\
Maximum generation length & 1,024 \\
Generation temperature & 0.99 \\
Value model & enabled \\
Tactic heartbeats & $10^8$ \\
Tactic timeout & 10,000 ms \\
\midrule
$c_{\mathrm{init}}$ & 0.001 \\
$c_{\mathrm{base}}$ & 3,200 \\
$\gamma$ & 0.99 \\
Prior temperature & 150 \\
Progressive-sampling $C$ & 0.05 \\
Progressive-sampling $\alpha$ & 0.6 \\
\bottomrule
\end{tabular}
\end{table}

The MCTS-specific hyperparameters in
Table~\ref{tab:search-hyperparameters} instantiate the progressive-sampling rule
in Eq.~(\ref{eq:progressive-sampling}) and the PUCT rule in
Eq.~(\ref{eq:puct}), with the exploration coefficient defined in
Eq.~(\ref{eq:puct-exploration}). The 65,536-node cap is set above the
8,192-step simulation budget and primarily serves as a safety bound on tree
growth.

Table~\ref{tab:training-config} reports the main training hyperparameters not
detailed in the main text. The MCTS simulation budget doubles across successive
RL stages, increasing from 512 simulations per search in RL1 to 8,192 in RL5.

\begin{table}[!htbp]
\caption{Training hyperparameters for Reaper-1.}
\label{tab:training-config}
\centering
\small
\begin{tabular}{ll}
\toprule
\textbf{Setting} & \textbf{Configuration} \\
\midrule
CPT global batch size & 1,024 \\
CPT learning rate & $3\times10^{-4} \rightarrow 3\times10^{-5}$ \\
CPT schedule & Cosine decay \\
\midrule
RL learning rate & $2\times10^{-5} \rightarrow 5\times10^{-7}$ \\
RL policy/value loss-weight ratio $\lambda_P : \lambda_V$ & $1 : 0.3 \rightarrow 1 : 0.05$ \\
RL1 MCTS budget & 512 simulations/search \\
RL2 MCTS budget & 1,024 simulations/search \\
RL3 MCTS budget & 2,048 simulations/search \\
RL4 MCTS budget & 4,096 simulations/search \\
RL5 MCTS budget & 8,192 simulations/search (B00--B24) \\
RL5 passes per problem & 8 (B00--B24) \\
\bottomrule
\end{tabular}
\end{table}

\section{Premise Selector Data and Training}
\label{app:ps-data}

Our premise-selector dataset follows the state--positive-premise construction used by
LeanSearch-PS~\citep{realprover}, with the Mathlib database updated to the
v4.28.0-rc1 snapshot. jixia and Scythe first extracts proof-state transitions and
their global references. For each transition, we retain a pair $(s,t^+)$ only
when a referenced declaration is present in the filtered Mathlib corpus and its
last identifier component occurs as a standalone token in the original tactic.
This keeps the label tied to an explicit user-facing theorem reference and allows
one query to have multiple positives.

The filtering is deliberately conservative. We keep ordinary nonempty proof
states containing a goal marker and remove wrapper/parser nodes, punctuation-only
nodes, placeholders, and \texttt{calc} aggregation nodes. The corpus filter keeps
declarations in Mathlib with a formal statement, while excluding internal or
implementation-oriented modules and names (including tactic, parser, elaborator,
private, generated, and unsafe-recursion entries) and selected constructor,
recursor, and opaque declarations. Exact query--positive pairs are then
deduplicated, and the train/validation split is made by query so that all
positives for one state remain in the same split.

For reference, the recorded construction starts from 1,024,367 extracted records,
removes 13,749 empty states, 7,650 nonempty states without a goal marker, and 50
\texttt{calc} aggregation records, and leaves 1,002,918 records before pair
deduplication. Deduplication removes 105,504 repeated query--positive pairs,
leaving 897,414 unique pairs. With seed 42, the query-grouped split contains
896,902 training pairs and 512 validation pairs from 226 validation queries.

For a query embedding $q_i$, let $d_{i,y_i}$ be its annotated positive and let
$s_{ij}=\langle q_i,d_j\rangle$ be the in-batch query--document similarity.
The selector is trained with the masked InfoNCE objective~\citep{infonce}

\begin{equation}
\mathcal{L}_{\mathrm{PS}}
=
-\frac{1}{B}\sum_{i=1}^{B}
\log
\frac{\exp(\widetilde{s}_{i,y_i}/\tau)}
{\sum_j\exp(\widetilde{s}_{ij}/\tau)},
\qquad
\widetilde{s}_{ij}=
\begin{cases}
-\infty, & s_{ij}>s_{i,y_i}+\delta,\\
s_{ij}, & \text{otherwise},
\end{cases}
\label{eq:premise-selector-loss}
\end{equation}

where $\tau$ and $\delta$ control the score-based fake-negative mask. Candidates
that are substantially more similar than the recorded positive are removed
from the denominator in Eq.~(\ref{eq:premise-selector-loss}). The objective
remains one-directional (query to premise); the mask is a conservative way to
reduce the penalty from one-to-many labels and semantically related in-batch
candidates.

The selector is trained with full-parameter ms-swift embedding SFT using
InfoNCE, batch negatives, a temperature of 0.1, and the fake-negative mask in
Eq.~(\ref{eq:premise-selector-loss}). The reported selector uses
Qwen3-Embedding-8B, three epochs,
learning rate $5\times10^{-5}$, maximum sequence length 512, bf16 precision,
per-device batch size 8, eight data-parallel workers, and dropped incomplete
training batches. No hard-negative mining or auxiliary symmetric query--query
objective is used in this run; the selector is trained only to rank premises for
the current proof-state query.

\section{Detailed Analysis of Reinforcement-Learning Dynamics}
\label{app:rl-effect-study}

We study not only whether RL solves more theorems, but also how it changes the reasoning during MCTS search. The analysis has two complementary parts.
First, we compare SFT, RL4, and RL5 on the same 244 miniF2F-test problems under 32 independent searches and 8,192 MCTS simulations per search.
Second, we follow 25 consecutive online-RL rollout batches to characterize when
the observed style changes emerge. The paired comparison controls the theorem
set and is therefore our primary evidence for checkpoint-level changes; the
batch sequence is observational because its theorem sets evolve over training. 

\subsection{Measurement Protocol}

A theorem is counted as solved only when at least one saved trajectory contains
a nonempty proof that passes final replay in Lean. For each solved theorem, its
\emph{canonical proof} is the verified proof from the earliest successful pass,
breaking ties by the smaller number of top-level actions.

Proof-style comparisons use the 151-problem \emph{paired common core} solved by
all three checkpoints. This prevents changes in theorem difficulty from being
mistaken for changes in style. A proof step is one action submitted by MCTS to
Lean. Such an action can contain the \texttt{<;>} combinator, multiple lines, or
a nested \texttt{by} block, and is therefore not necessarily an atomic tactic.
We consequently report action length and internal tactic density alongside
top-level step count. 

For search-wide cost, we additionally stream all completed
\texttt{tactic\_eval} records with valid start and stop timestamps. Here,
``accepted'' means that the Lean evaluator returned no error. Exact counts, means, acceptance rates, and structural shares are computed over all calls. Reported quantiles use deterministic uniform
reservoir samples. Unless stated otherwise, confidence intervals are
problem-clustered bootstrap intervals, solved-set transitions use the exact
McNemar test, and paired continuous outcomes use a two-sided Wilcoxon
signed-rank test on $\log(1+x)$.

\subsection{Capability Expansion and Successful-Search Efficiency}

Table~\ref{tab:rl-checkpoint-efficiency} separates coverage on all 244 problems
from proof and search statistics on the 151-problem common core. The dominant
capability gain occurs by RL4: pass@32 increases from 62.3\% to
78.3\%, with 40 newly solved problems and one loss relative to SFT (exact
McNemar $p=3.82\times10^{-11}$). RL5 reaches 79.9\% and is a strict superset of
the RL4 solved set, adding four problems without a loss.

\begin{table}[!htbp]
\caption{Coverage and successful-search efficiency across checkpoints.
Coverage uses all 244 miniF2F-test problems; the remaining rows use one
canonical proof for each of the 151 problems solved by all three checkpoints.}
\label{tab:rl-checkpoint-efficiency}
\centering
\small
\begin{tabular}{lrrr}
\toprule
\textbf{Metric} & \textbf{SFT} & \textbf{RL4} & \textbf{RL5} \\
\midrule
Pass@32 & 152 (62.3\%) & 191 (78.3\%) & 195 (79.9\%) \\
Proof steps, mean & 21.7 & 11.2 & 5.2 \\
Proof steps, median [IQR] & 7 [2, 18] & 2 [1, 8.5] & 2 [1, 3.5] \\
One-step proofs & 23.2\% & 49.0\% & 49.0\% \\
Tree nodes, mean & 2,368 & 162 & 119 \\
Tree nodes, median [IQR] & 901 [10, 3,891] & 3 [2, 18] & 3 [2, 7] \\
Root visits, median & 922 & 2 & 2 \\
\bottomrule
\end{tabular}
\end{table}

\begin{figure}[!htbp]
\centering
\includegraphics[width=0.98\linewidth]
    {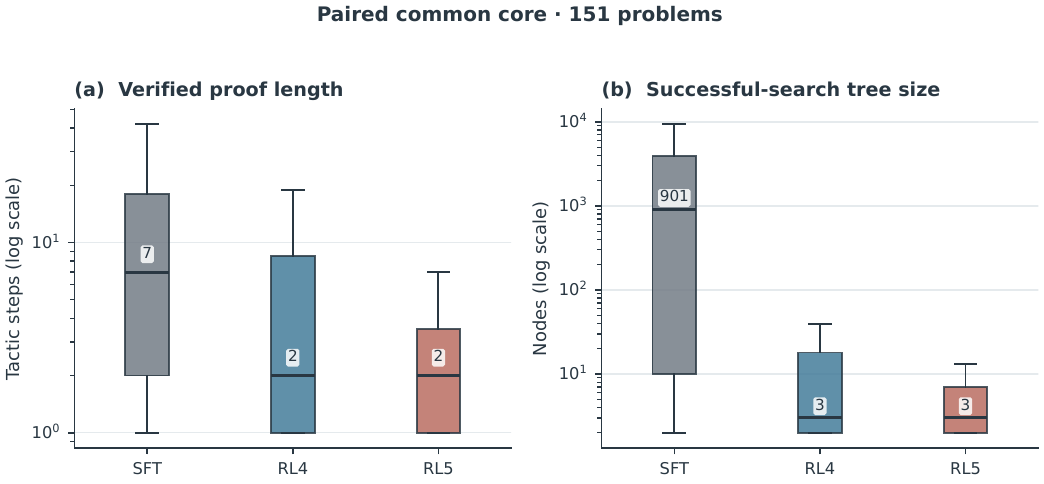}
\caption{Verified proof length and successful-search tree size on the paired
151-problem common core. Each checkpoint contributes one earliest-success
canonical proof per theorem. Boxes show the interquartile range and the
annotated line is the median; both axes are logarithmic.}
\label{fig:rl-common-core}
\end{figure}

Figure~\ref{fig:rl-common-core} shows that the gain is not only in final
coverage. From SFT to RL4, mean proof length falls by 10.44 top-level actions
(95\% bootstrap CI $[-20.93,-2.15]$), and the mean successful tree shrinks by
2,206 nodes ($[-2,731,-1,723]$). Both paired changes are significant
($p<10^{-8}$ and $p<10^{-20}$, respectively). RL5 further reduces the mean
proof length and tree size, but their medians remain at two actions and three
nodes. Thus, the later-stage effect is concentrated in the long tail: among the
191 theorems shared by RL4 and RL5, 81 proofs become shorter, 78 are unchanged,
and 32 become longer, while the mean step difference is $-11.88$
($[-18.01,-6.80]$). Selecting each theorem's shortest verified proof instead of
its earliest one preserves the same ordering (11.8, 7.8, and 3.9 mean actions
for SFT, RL4, and RL5).

The tree reduction is especially important for interpreting the proof-length
result. It indicates that RL often causes a successful continuation to be found
after expanding fewer search states, rather than merely shortening the emitted
text.

\subsection{From Solver Calls to Structured Macro-Actions}

\begin{figure}[!htbp]
\centering
\includegraphics[width=0.98\linewidth]
    {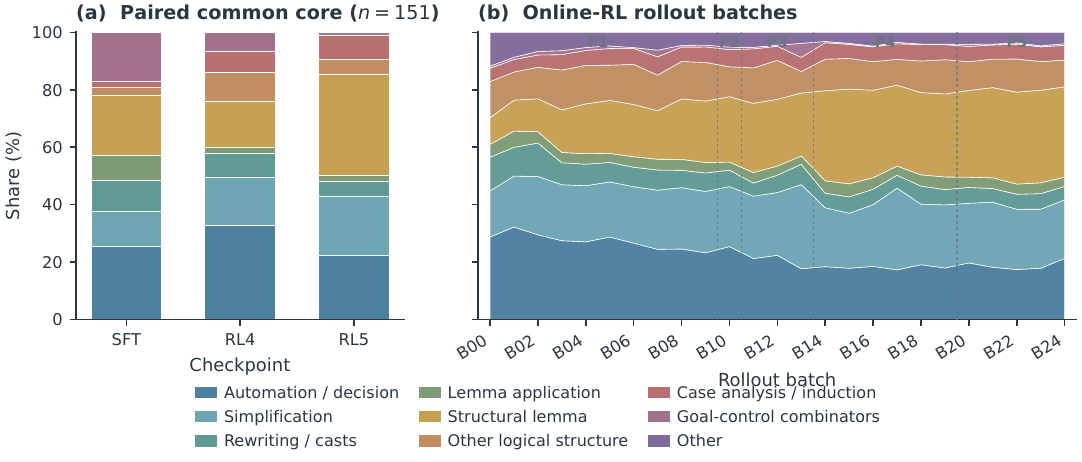}
\caption{Outer-head tactic composition with a shared taxonomy and legend.
(a) Action-weighted shares on the paired common core across checkpoints.
(b) Shares across online-RL rollout batches.}
\label{fig:rl-tactic-mix}
\end{figure}

The successful proofs exhibit a two-stage style transition. We classify an
action by its outermost tactic head. The \emph{structural lemma} category
contains \texttt{have}, \texttt{suffices}, \texttt{replace}, \texttt{let},
\texttt{haveI}, \texttt{wlog}, and \texttt{set}; solvers nested inside these
actions are not counted as outer automation. Figure~\ref{fig:rl-tactic-mix}
therefore describes the organization of top-level search actions, while
Table~\ref{tab:rl-action-style} measures their internal complexity.

\begin{table}[!htbp]
\caption{Granularity of canonical-proof actions on the paired common core.
``Identifiers/action'' counts tactic identifiers lexically throughout each
action, including nested blocks.}
\label{tab:rl-action-style}
\centering
\small
\begin{tabular}{lrrr}
\toprule
\textbf{Metric} & \textbf{SFT} & \textbf{RL4} & \textbf{RL5} \\
\midrule
Action characters, median & 19 & 23 & 66 \\
Tactic identifiers/action & 1.38 & 1.38 & 2.80 \\
Contains \texttt{<;>} & 2.9\% & 4.5\% & 23.9\% \\
Multiline action & 7.0\% & 3.5\% & 24.7\% \\
Contains nested \texttt{:= by} & 1.5\% & 3.8\% & 17.5\% \\
Explicit retrieved-premise use & 6.4\% & 7.8\% & 19.5\% \\
\bottomrule
\end{tabular}
\end{table}

The transition from SFT to RL4 is primarily a shift toward direct,
high-leverage search actions.
Outer automation rises from 25.5\% to 32.9\%, case analysis and induction from
2.1\% to 7.3\%, and goal-control combinators fall from 16.8\% to 6.5\%. This
is consistent with RL4 learning to invoke arithmetic solvers or split the
relevant cases earlier, thereby collapsing the search tree.

The transition from RL4 to RL5 changes the abstraction boundary.
Structural-lemma actions rise
from 15.9\% to 35.3\%, while outer automation falls to 22.2\%. The decline in
outer automation does not mean that solvers disappear: per 100 top-level
actions, lexical occurrences of \texttt{norm\_num} increase from 23.4 to 48.1,
\texttt{simp} from 13.8 to 30.2, and \texttt{have} from 16.1 to 41.9. Solvers
are increasingly embedded inside local facts, branches, and compound tactic
programs. The top-level action count therefore compresses at the same time that
each action becomes longer and internally denser.

This distinction matters conceptually. Later RL learns a more programmatic
proof interface for MCTS---one action can introduce a lemma, dispatch several
subgoals, normalize expressions, and invoke one or more solvers. We refer to
this as \emph{search-interaction compression}, not a direct reduction in
reasoning depth. The larger premise-use share further suggests that RL5 more
often integrates retrieved library facts into these structured programs.

\subsection{Where the Computation Moves}

\begin{figure}[!htbp]
\centering
\includegraphics[width=0.98\linewidth]
    {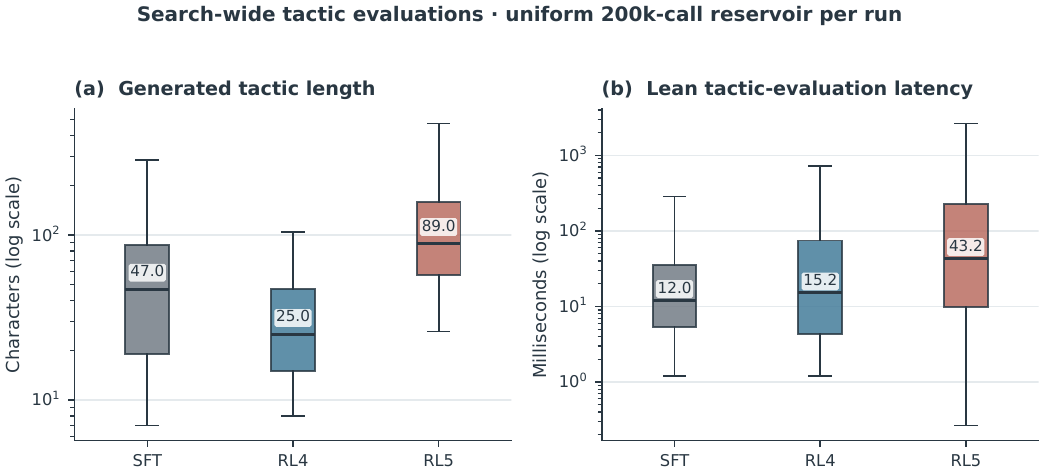}
\caption{Search-wide distributions of generated tactic length and Lean
evaluation latency. Boxes show the interquartile range, whiskers show the 5th
and 95th percentiles, and annotations mark medians. Each run contributes a
uniform 200,000-call reservoir; axes are logarithmic.}
\label{fig:rl-eval-distributions}
\end{figure}

To determine whether fewer search interactions translate into cheaper proof
search, we analyze 48.57 million completed tactic evaluations across the three
runs. Table~\ref{tab:rl-eval-cost} and
Figure~\ref{fig:rl-eval-distributions} reveal a non-monotone trajectory. RL4
generates shorter, more frequently accepted candidates than SFT, but its use of
computationally heavy solvers makes the median call slightly slower. RL5 then
shifts probability mass toward long macro-actions: the median candidate grows
to 89 characters, median evaluator latency reaches 43.2 ms, and acceptance
falls to 35.0\%.

\begin{table}[!htbp]
\caption{Search-wide Lean tactic evaluations. Counts and rates are exact over
all completed calls; quantiles use a deterministic 200,000-call uniform
reservoir per run. Summed duration is overlapping service time, not wall-clock
runtime.}
\label{tab:rl-eval-cost}
\centering
\small
\begin{tabular}{lrrr}
\toprule
\textbf{Metric} & \textbf{SFT} & \textbf{RL4} & \textbf{RL5} \\
\midrule
Completed evaluator calls & 24.57M & 18.01M & 5.99M \\
Tactic chars, median [IQR] & 47 [19, 87] & 25 [15, 47] & 89 [57, 159] \\
Latency ms, median [IQR] & 12.0 [5.4, 35.0] & 15.2 [4.4, 74.6] & 43.2 [9.7, 229.2] \\
Evaluator accepted & 58.1\% & 61.3\% & 35.0\% \\
Calls exceeding 1 second & 1.85\% & 3.84\% & 9.73\% \\
Calls/problem, median & 29,512 & 992.5 & 257 \\
Summed evaluator duration & 699 h & 1,067 h & 805 h \\
\bottomrule
\end{tabular}
\end{table}

Much of RL5's raw latency and error penalty is a composition effect rather than
a uniform degradation. In the 64--127-character bin, for example, RL4 and RL5
have nearly identical acceptance (33.0\% versus 34.2\%) and accepted-call
median latency (46.6 versus 47.1 ms). Nevertheless, longer programs create
more opportunities for parse, elaboration, or tactic failure, and RL5 places
far more mass in the long bins. Error calls are also slower than accepted calls
at every checkpoint.

The number of completed calls per problem drops dramatically, but total
evaluator work does not fall in proportion. From RL4 to RL5, call count falls
by 66.7\%, whereas summed evaluator duration falls by only 24.6\%. From SFT to
RL4, summed evaluator duration even rises despite fewer calls. Later RL thus
reallocates computation from tree expansion to execution inside each action.
Optimizing only proof success or top-level trajectory length can favor this
exchange; a deployment-oriented objective may additionally need to regularize
generated tokens, evaluator errors, latency, or the expanded cost of tactic
combinators.

\subsection{Longitudinal Dynamics over 25 Online-RL Batches}

The checkpoint comparison gives a controlled view of three endpoints. To
resolve the transition in more detail, we analyze the 25 consecutive rollout
batches (B00--B24), comprising 25,069 problem--batch instances, 56,959 verified
successful attempts, and 345.6 million completed tactic evaluations. Adjacent batches use different theorem sets, so
the resulting trends describe the joint training-and-rollout process rather
than a pure checkpoint treatment effect.

\begin{figure}[!htbp]
\centering
\includegraphics[width=0.98\linewidth]
    {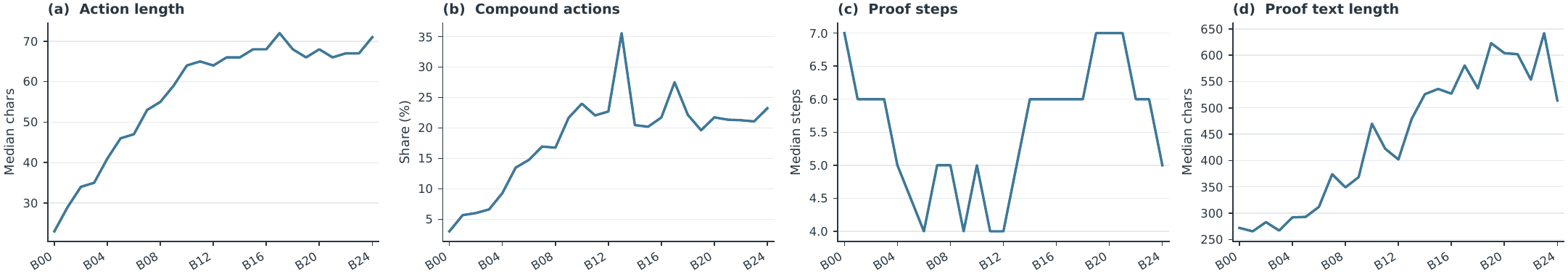}
\caption{Canonical-proof granularity across online-RL batches: (a) median
action length, (b) compound-action share, (c) median top-level proof steps,
and (d) median proof-text length.}
\label{fig:rl-online-style}
\end{figure}

The most stable longitudinal signal is the growing size of a top-level action.
From B00 to B24, median action length increases from 23 to 71 characters, the
share containing \texttt{<;>} from 3.0\% to 23.2\%, and median proof text from
272 to 514 characters, while median top-level proof length changes only from
seven to five actions. Across canonical proofs, structural-lemma heads rise
from 9.3\% in B00 to 31.4\% in B24. Their
batch-wise share has Spearman $\rho=0.929$ ($p=2.00\times10^{-11}$), and almost
all of the increase comes from \texttt{have}. Lexical solver use rises at the
same time. This reproduces the endpoint result: proof programs become more
locally structured, not merely more reliant on automation.

\begin{figure}[!htbp]
\centering
\includegraphics[width=0.98\linewidth]
    {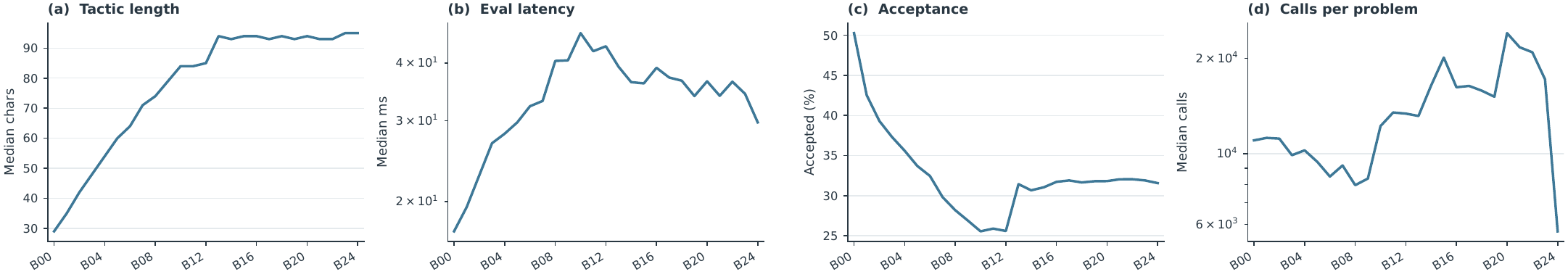}
\caption{Search-wide tactic-evaluation trends across online-RL batches:
(a) median tactic length, (b) median evaluation latency, (c) acceptance rate,
and (d) median evaluation calls per problem.}
\label{fig:rl-online-eval}
\end{figure}

Search-wide evaluation reveals an early transition followed by stabilization. During
B00--B09, median evaluated tactic length grows from 29 to 79 characters,
median latency from 17.2 to 40.6 ms, and acceptance falls from 50.3\% to
26.9\%. B10--B12 forms the lowest-acceptance regime: median actions are 84--85
characters and acceptance is only 25.6--25.9\%. At B13, median action length
reaches 94 characters, yet acceptance jumps to 31.4\% and median latency falls
to 39.3 ms. Thereafter, length remains near 93--95 characters, acceptance near
31--32\%, and latency mostly between 34 and 39 ms.

A coarse adjustment using a common action-length-bin distribution gives the
same conclusion. Comparing B00--B12 with B13--B24, length-standardized
acceptance increases from 28.9\% to 34.9\% ($p=2.49\times10^{-5}$ across
batches), while standardized mean evaluator duration falls from 559.9 to
442.6 ms ($p=1.01\times10^{-4}$). Later RL therefore appears to improve the
executability of an already learned macro-action representation rather than
continuing to enlarge it indefinitely. This improvement does not by itself
imply lower end-to-end compute: longer individual tactic evaluations can offset
reductions in the number of search interactions even under a fixed search
budget.

\section{Qualitative Case Study: Proof-Style Evolution}
\label{app:case-proof-evolution}

The aggregate analysis in Appendix~\ref{app:rl-effect-study} shows that later
RL compresses long successful trajectories while increasingly packaging
several Lean operations into one top-level action. The value objective in
Reaper-1 assigns a unit cost to each tactic-level search action and trains the
value head toward the remaining cost of successful continuations. The
controlled ablation in Table~\ref{tab:value_ablation} further shows that this
objective reduces average successful-proof length while yielding similar overall coverage. We now examine a representative theorem to
show what the corresponding change looks like inside a complete proof.

Listings~\ref{lst:proof-rl4} and~\ref{lst:proof-rl5} show complete Lean proofs
produced by the RL4 and RL5 checkpoints for the same miniF2F theorem. Both
prove the identity
\[
\sum_{k < n} k^3
=
\left(\sum_{k < n} k\right)^2,
\]
but they reach the result through substantially different proof trajectories.

\paragraph{RL4 proof.}
The RL4 checkpoint discovers a valid proof, but its trajectory is dominated by
repeated local normalization and repair. After the initial induction, the proof
repeatedly invokes \texttt{simp}, \texttt{ring}, \texttt{ring\_nf},
\texttt{rw}, and automation, introduces a second induction, and revisits
closely related algebraic states multiple times.

\begin{lstlisting}[style=leancode,
caption={Complete proof generated by the RL4 checkpoint.},
label={lst:proof-rl4}]
import Mathlib 
import Aesop 

set_option maxHeartbeats 0 

open BigOperators Real Nat Topology Rat

theorem induction_sumkexp3eqsumksq_rl4 (n : ℕ) :
    (∑ k ∈ Finset.range n, k ^ 3) = (∑ k ∈ Finset.range n, k) ^ 2 := by
  induction n
  (*@\leanbullet@*) 
    simp [Finset.sum_range_succ]
  (*@\leanbullet@*)
    simp [Finset.range_succ, Finset.sum_insert, *]
    ring
    abel
    simp_all [pow_succ]
    ring_nf
    simp [pow_succ]
    simp [mul_assoc]
    clear * -
    induction' ‹ℕ› with n hn
    (*@\leanbullet@*)
      rfl
    (*@\leanbullet@*)
      simp [Finset.sum_range_succ]
      ring_nf
      rw [hn]
      ring_nf
      abel_nf
      rw [add_comm]
      cases n
      (*@\leanbullet@*)
        rfl
      (*@\leanbullet@*)
        ring
        simp [add_comm, add_left_comm, add_assoc,
          mul_comm, mul_left_comm, mul_assoc] at *
        ring
        simp [Finset.sum_range_succ] at *
        ring_nf at hn
        simp [add_assoc] at *
        simp [add_assoc, add_comm, add_left_comm] at *
        simp [Finset.sum_range_succ] at *
        ring
        rw [Finset.sum_range_id]
        ring_nf at hn ⊢
        simp [mul_add, mul_comm, mul_left_comm] at *
        ring
        simp [add_assoc, add_comm, add_left_comm] at *
        ring
        simp [Nat.mul_sub_left_distrib,
          Nat.mul_sub_right_distrib, Nat.mul_add,
          Nat.mul_comm, Nat.mul_left_comm] at hn ⊢
        ring_nf
        ring_nf at *
        simp [mul_assoc, mul_comm, mul_left_comm] at hn ⊢
        ring
        symm
        ring_nf at *
        have := hn
        aesop
        cases n
        (*@\leanbullet@*)
          norm_num [this]
        (*@\leanbullet@*)
          ring
          rw [add_comm]
          have this := this
          ring
          ring_nf at *
          aesop
          cases n
          (*@\leanbullet@*)
            grind
          (*@\leanbullet@*)
            ring
            ring_nf at this
            revert this
            repeat rw [Finset.sum_range_id]
            grind [mul_succ, mul_add, add_mul, add_assoc,
              add_comm, add_left_comm, Nat.pow_succ,
              Nat.mul_sub_left_distrib,
              Nat.mul_sub_right_distrib,
              Nat.div_eq_of_lt, Nat.succ_mul]
\end{lstlisting}

The proof has a strongly exploratory character: the model repeatedly changes
the algebraic normal form and applies automation.
In particular, several rounds of \texttt{ring}/\texttt{ring\_nf} and
\texttt{simp} are interleaved with nested case analysis and another induction.

\paragraph{RL5 proof.}
By RL5, the proof is considerably more concentrated. The checkpoint starts from
the same induction structure, immediately incorporates the induction hypothesis,
rewrites the square and the closed form of the finite sum, and reduces the
remaining obligation to a targeted parity split followed by arithmetic
automation.

\begin{lstlisting}[style=leancode,
caption={Complete proof generated by the RL5 checkpoint for the same theorem.},
label={lst:proof-rl5}]
theorem induction_sumkexp3eqsumksq_rl5 (n : ℕ) :
    (∑ k ∈ Finset.range n, k ^ 3) = (∑ k ∈ Finset.range n, k) ^ 2 := by
  induction n with
    | zero => ?_
    | succ n ih => ?_
  (*@\leanbullet@*)
    simp_all
  (*@\leanbullet@*)
    simp [Finset.sum_range_succ, ih]
    rw [add_sq]
    rw [Finset.sum_range_id]
    cases n
    (*@\leanbullet@*)
      simp at ih ⊢
    (*@\leanbullet@*)
      ring_nf
      simp [add_assoc, add_comm, add_left_comm] at ih ⊢
      ring_nf at *
      simp [Nat.add_assoc, Nat.add_comm,
        Nat.add_left_comm] at *
      aesop
        (add simp [Nat.pow_two, Nat.mul_div_cancel_left])
      rcases n.even_or_odd' with (*@$\langle$@*)m, rfl | rfl(*@$\rangle$@*) <;>
        grind [Nat.mul_div_cancel']
\end{lstlisting}

The difference is not merely textual brevity. The RL5 proof follows a much
clearer global progression:
\emph{induction $\rightarrow$ use of the induction hypothesis $\rightarrow$
algebraic normalization $\rightarrow$ targeted parity split $\rightarrow$
arithmetic closure}. In contrast, the RL4 proof spends many more search actions
on repeated normalization, nested case analysis, and local recovery.

This behavior is consistent with the intended effect of Reaper-1's
cost-sensitive value-guided search ablation in Table~\ref{tab:value_ablation}. Because each tactic-level transition incurs a
unit cost, lower-cost successful continuations receive more favorable value
targets, creating pressure toward proof trajectories that reach solved states
through fewer search decisions. This example further illustrates what the corresponding change can look like
at the level of an individual proof.

Shorter continuations can also be advantageous to MCTS because they reduce the
effective depth at which a successful leaf must be discovered, allowing a
fixed simulation budget to concentrate on fewer proof decisions. We view this
as one mechanism that may contribute to the continued improvement on harder
problems in later RL stages. Moreover, the final RL5 proof also illustrates a limitation of
our current cost definition: the use of
\texttt{<;>} packages multiple internal tactic applications into one
search-level action. Thus, the learned preference for shorter trajectories
captures genuine removal of redundant search in this example, but may also
encourage syntactic compression of several Lean operations into a single
action, as discussed in Section~\ref{sec:limitations}.

\section{Tactic Normalization}
\label{app:tactic-normalization}

Lean's structural tactics expose points at which the proof state branches or
changes shape. Tactics such as \texttt{have}, \texttt{suffices}, and
\texttt{calc} can therefore provide valuable training signals, including the
decomposition choices that precede the final proof steps. During ordinary
elaboration, however, these tactics expect their proof bodies to be supplied
as complete tactic blocks. A standalone action such as
\texttt{have ... := by} leaves a nested goal unfinished and cannot be replayed
as a valid tactic transition.

We normalize such actions by replacing each nested \texttt{by} block with a
placeholder and exposing the resulting subgoals with Lean's bullet syntax
(``\texttt{·}''). The normalized form is a valid tactic sequence: the outer
structural tactic remains visible as one search action, while each generated
subgoal is solved by a separate bullet. This preserves the original goal
decomposition and makes the transition available to training and analysis.

The following pairs show the transformation for three representative
structural tactics. In each pair, the first listing is the original nested
form and the second is the bullet-normalized form.

\paragraph{The \texttt{have} tactic.}
Here we replace the nested proof of \texttt{key} with \texttt{?\_} and expose
the continuation and the new subgoal as two bullet goals. Listings~\ref{lst:norm-have-original}
and~\ref{lst:norm-have} show the original and normalized forms.

\begin{lstlisting}[style=leancode,
caption={Original nested \texttt{have} proof.},
label={lst:norm-have-original}]
theorem coe_eq_inv_of_sq_eq_one
    (hφ : FixedPointFree φ) (h2 : φ^[2] = _root_.id) :
    ⇑φ = (·⁻¹) := by
  ext g
  have key : g * φ g = 1 := by
    simpa [List.range_succ] using hφ.prod_pow_eq_one h2 g
  rwa [← inv_eq_iff_mul_eq_one, eq_comm] at key
\end{lstlisting}

\begin{lstlisting}[style=leancode,
caption={Bullet-normalized \texttt{have} proof.},
label={lst:norm-have}]
theorem coe_eq_inv_of_sq_eq_one
    (hφ : FixedPointFree φ) (h2 : φ^[2] = _root_.id) :
    ⇑φ = (·⁻¹) := by
  ext g
  have key : g * φ g = 1 := ?_
  · rwa [← inv_eq_iff_mul_eq_one, eq_comm] at key
  · simpa [List.range_succ] using hφ.prod_pow_eq_one h2 g
\end{lstlisting}

\paragraph{The \texttt{suffices} tactic.}
The nested conclusion is replaced by \texttt{from ?\_}; the two bullets then
prove the conclusion and the auxiliary equivalence introduced by
\texttt{suffices}, respectively. Listings~\ref{lst:norm-suffices-original}
and~\ref{lst:norm-suffices} show the corresponding pair.

\begin{lstlisting}[style=leancode,
caption={Original nested \texttt{suffices} proof.},
label={lst:norm-suffices-original}]
/-- Conjugacy classes form a partition of G, stated in terms of cardinality. -/
theorem sum_conjClasses_card_eq_card [Fintype <| ConjClasses G] [Fintype G]
    [∀ x : ConjClasses G, Fintype x.carrier] :
    ∑ x : ConjClasses G, x.carrier.toFinset.card = Fintype.card G := by
  suffices (Σ x : ConjClasses G, x.carrier) ≃ G by
    simpa using (Fintype.card_congr this)
  simpa [carrier_eq_preimage_mk] using Equiv.sigmaFiberEquiv ConjClasses.mk
\end{lstlisting}

\begin{lstlisting}[style=leancode,
caption={Bullet-normalized \texttt{suffices} proof.},
label={lst:norm-suffices}]
/-- Conjugacy classes form a partition of G, stated in terms of cardinality. -/
theorem sum_conjClasses_card_eq_card [Fintype <| ConjClasses G] [Fintype G]
    [∀ x : ConjClasses G, Fintype x.carrier] :
    ∑ x : ConjClasses G, x.carrier.toFinset.card = Fintype.card G := by
      suffices (Σ x : ConjClasses G, x.carrier) ≃ G from ?_
      · simpa using (Fintype.card_congr this)
      · simpa [carrier_eq_preimage_mk] using Equiv.sigmaFiberEquiv ConjClasses.mk
\end{lstlisting}

\paragraph{The \texttt{calc} tactic.}
Each nested proof attached to a calculation step becomes a placeholder, and
the four resulting equality goals are discharged by bullets after the
calculation block. Listings~\ref{lst:norm-calc-original} and
\ref{lst:norm-calc} show this conversion.

\begin{lstlisting}[style=leancode,
caption={Original nested \texttt{calc} proof.},
label={lst:norm-calc-original}]
/-- In a nontrivial monoid with zero, zero and one are different. -/
instance NeZero.one : NeZero (1 : M₀) := ⟨by
  intro h
  rcases exists_pair_ne M₀ with ⟨x, y, hx⟩
  apply hx
  calc
    x = 1 * x := by rw [one_mul]
    _ = 0 := by rw [h, zero_mul]
    _ = 1 * y := by rw [h, zero_mul]
    _ = y := by rw [one_mul]⟩
\end{lstlisting}

\begin{lstlisting}[style=leancode,
caption={Bullet-normalized \texttt{calc} proof.},
label={lst:norm-calc}]
/-- In a nontrivial monoid with zero, zero and one are different. -/
instance NeZero.one : NeZero (1 : M₀) := ⟨by
  intro h
  rcases exists_pair_ne M₀ with ⟨x, y, hx⟩
  apply hx
  calc
    x = 1 * x := ?_
    _ = 0 := ?_
    _ = 1 * y := ?_
    _ = y := ?_
  · rw [one_mul]
  · rw [h, zero_mul]
  · rw [h, zero_mul]
  · rw [one_mul]⟩
\end{lstlisting}

The same transformation is applied to the other structural tactics used by
the rollout pipeline, including \texttt{cases}, \texttt{induction},
\texttt{obtain}, \texttt{show}, \texttt{haveI}, \texttt{let}, and
\texttt{letI}. The normalized representation keeps these state transitions
explicit while remaining executable by Lean's elaborator.

\section{From Lean Source Code to MCTS Trees}
\label{app:elab-to-rollout}

To support SFT on existing Lean codebases, we use Jixia to generate elaboration traces for the source code, which include the proof states before and after each
tactic, together with the tactic's source span and the constants referenced in
its elaborated subtree. These traces describe successful proofs, but
they are not yet represented in the raw MCTS-tree schema used by Scythe-rl. In particular, when a tactic replaces one goal
with several goals, the resulting proof obligation must be represented by an AND node.

We traverse jixia's elaboration tree to collect executable tactic events, then
reconstruct proof transitions from their recorded input and output states and
source ranges. This is necessary because the elaboration tree contains other nodes such as 
parser wrappers, compound tactics, and nested local proof blocks, which are irrelevant in our state-action model.

In the resulting tree, each proof state is represented by an OR node with one child, where the outgoing edge is the tactic used on that state. If the tactic produces exactly one remaining
goal, the edge leads directly to the OR node for that goal. If it produces
multiple goals, the edge instead leads to an AND node, which then leads to the OR nodes for this set of goals.

The result is then serialized using the same MCTS-tree schema as proof-search rollouts. Each real tactic edge records the constants found in its elaborated
subtree in the \texttt{premise} field. The format is fully compatible with MCTS trees generated by proof search. The converted tree
can therefore enter the same replay and training pipeline as a tree produced
by proof search, allowing existing Lean source repositories, such as Mathlib, to be freely mixed alongside search-generated rollouts.
\end{document}